\pdfoutput=1

\documentclass[11pt, table]{article}

\usepackage[final]{acl}
\usepackage{times}
\usepackage{latexsym}

\usepackage[T1]{fontenc}

\usepackage[utf8]{inputenc}

\usepackage{microtype}

\usepackage{inconsolata}

\usepackage{graphicx}
\usepackage{xcolor}
\usepackage{booktabs}
\usepackage{amsmath}
\usepackage{amsfonts}
\usepackage{amssymb}
\usepackage[ruled,vlined,linesnumbered]{algorithm2e}

\SetCommentSty{mycommfont}
\usepackage{multirow}
\usepackage{makecell}
\usepackage{siunitx}
\usepackage{natbib}
\usepackage{graphicx}
\usepackage{subcaption}
\usepackage{ulem}
\usepackage[export]{adjustbox}
\definecolor{improve}{RGB}{34,139,34}
\definecolor{degrade}{RGB}{255,0,0}
\usepackage{tcolorbox}
\usepackage{enumitem}

\newcommand{\red}[1]{{\color{red}{#1}}}
\newcommand{\green}[1]{{\color{improve}{#1}}}

\usepackage{ragged2e}
\newtcolorbox{problemformat}{
    colback=white,
    colframe=black,
    arc=0pt,
    boxsep=5pt,
    left=2pt,
    right=2pt,
    top=1pt,
    bottom=1pt
}
\usepackage{float}

\title{\includegraphics[width=1em]{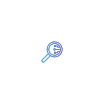}\ Large Language Models Have Intrinsic Meta-Cognition, \\but Need a Good Lens}

\author{
 \textbf{Ziyang Ma\textsuperscript{1*}},
 \textbf{Qingyue Yuan\textsuperscript{2*}},
 \textbf{Zhenglin Wang\textsuperscript{1}},
 \textbf{Deyu Zhou\textsuperscript{1\textdagger}
 }
\\
 \textsuperscript{1} School of Computer Science and Engineering, Key Laboratory of Computer Network\\ and Information Integration, Ministry of Education, Southeast University, China\\
 \textsuperscript{2} Department of Neurosurgery, Shanghai Tenth People's Hospital, School of \\ Clinical Medicine of Nanjing Medical University, China
\\
 {
   \texttt{\{mazy, zhenglin, d.zhou\}@seu.edu.cn \quad yuanqy007@stu.njmu.edu.cn}
 }
}

\begin{document}
\maketitle
\renewcommand{\thefootnote}{\fnsymbol{footnote}}
\footnotetext[1]{These authors contributed equally to this work.}
\footnotetext[2]{Corresponding Author.}
\renewcommand{\thefootnote}{\arabic{footnote}}

\begin{abstract}
Previous research has primarily focused on the cognitive error detection capabilities of Large Language Models (LLMs), often prompting them to analyze mistakes in reasoning chains. However, few studies have examined the meta-cognitive abilities of LLMs (e.g., their self-awareness of step errors), which are crucial for their reliability. While studies on LLM self-evaluation present some measures, such as perplexity, which can reflect the answer correctness and be viewed as the lens of meta-cognition, they lack step-level analysis and adaptation. This paper studies the evaluation of LLM meta-cognition using the current lenses and how to improve these lenses. Specifically, we propose AutoMeco, an Automated Meta-cognition Evaluation framework for benchmarking the existing lenses. Furthermore, a training-free Markovian Intrinsic Reward Adjustment strategy, MIRA, is proposed to boost current meta-cognition lenses. Experimental results on three mathematical reasoning datasets and three LLMs show the reasonableness of AutoMeco by comparing it with Best-of-N verification. Moreover, the meta-cognition ability of LLMs can be better evaluated using MIRA.\footnote{The code can be accessed via \url{https://github.com/Yann-Ma/AutoMeco}.}
\end{abstract}


\section{Introduction}
\label{introduction}

The reasoning ability of Large Language Models (LLMs) has improved tremendously with the emergence of Large Reasoning Models (LRMs) such as OpenAI-o1~\cite{jaech2024openai} and DeepSeek-R1~\cite{guo2025deepseek}. While evaluation on the cognitive capability of LLMs, such as reasoning outcome accuracy~\cite{lightman2023let,zeng2025mrgsmk}, presents the strength of LLMs, meta-cognition of these models that points to their consciousness of behavioral correctness is also important, especially for their reliability~\cite{zhou2024metacognitive, griot2025large, yan2025position}. 

\begin{figure}[t]
    \centering
    \includegraphics[width=\linewidth]{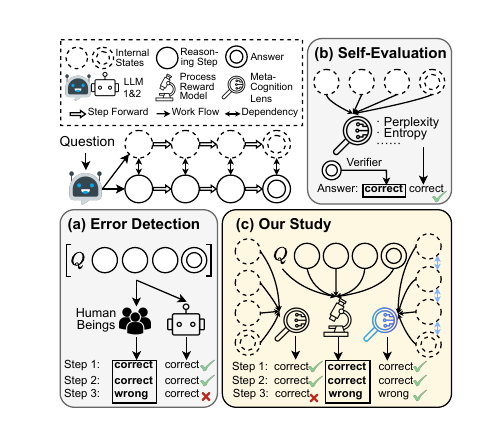}
    \caption{In reasoning tasks, error detection (a) focuses on LLMs' cognitive ability to analyze errors in reasoning steps. Self-evaluation (b) utilizes measures such as entropy as lenses to reflect self-awareness of answer rightness. Our work (c) studies the evaluation and improvement of the current lenses in reflecting LLM meta-cognition. \textbf{Bold} ``correct'' and ``wrong'' within boxes are ground truths of the answer or step correctness.}
    \label{fig:intro}
\end{figure}

In cognitive science, meta-cognition is the cognition beyond cognition, with subjective confidence of cognitive behaviors being the main indicator~\cite{matthews2018conscious,shea2019global}. Feeling of Rightness~\cite{thompson2011intuition} and Feeling of Error (FoE)~\cite{gangemi2015feelings} are two principal types of meta-cognition in reasoning tasks of human beings.
As shown in Figure~\ref{fig:intro}(a), previous research studies whether LLMs can detect error steps in responses generated by other models (e.g.,~\citealp{zeng2024mrben,tyen2024llms}), focusing on the cognitive error detection ability of LLMs. Besides, research on self-evaluation in Figure~\ref{introduction}(b) proves that internal states of LLMs can reflect the answer correctness through a trained linear classifier~\cite{zhang2025reasoning} or training-free measures such as perplexity~\cite{wang2025latent}, which can be viewed as the meta-cognition lenses. However, few of them concentrate on \textit{\uline{whether LLMs intrinsically have meta-cognition such as FoE during the reasoning process}}, which underlies the trustworthiness and self-improvement feasibility of LLMs.
Therefore, this paper studies the following two problems of LLM meta-cognition: (i) \textit{To what extent can LLM meta-cognition be observed through their internal states?} (ii) \textit{How to observe LLM meta-cognition based on their internal states without external sources more accurately?}

Evaluating and enhancing meta-cognition lenses for LLMs confronts two challenges rooted in the dataset and methodology. (1) \textbf{Data incompleteness} regarding the internal states of LLMs. To the best of our knowledge, existing error detection benchmarks contain no internal states of LLMs such as hidden states and probabilities~\cite{zeng2025mrgsmk, zeng2024mrben, xia2025evaluating}. This absence makes existing benchmarks unusable for the evaluation of meta-cognition lenses. (2) \textbf{Insufficient granularity} of existing lenses. While existing approaches mainly assess answer correctness \cite{si2022prompting,huang2023look,wang2025latent}, they ignore the sequential dependencies between steps and probably fail to provide stepwise signals.

To address these challenges, as depicted in Figure~\ref{fig:intro}(c): (1) We propose an \textbf{Auto}mated \textbf{Me}ta-\textbf{co}gnition Evaluation framework (\textbf{AutoMeco}) that realizes human-annotation-free benchmarking of LLM meta-cognition lenses. Our framework utilizes the Process Reward Model (PRM) as an annotator of step correctness. Furthermore, with the automatically annotated labels, the framework tests the lenses, such as perplexity, towards their step-rightness classification ability. 
(2) We propose a training-free \textbf{M}arkovian \textbf{I}ntrinsic \textbf{R}eward \textbf{A}djustment strategy (\textbf{MIRA}) that modifies the step rightness scores of the lenses based on Markov Decision Process (MDP) modeling and Q-value estimation. We model the step-level scoring as a Markov decision process, with dependencies between the reasoning steps. In the MDP, our adjustment strategy utilizes Q-value estimation to transmit the influences in reverse from the end.

The contributions of this paper are as follows:
\begin{itemize}
\setlength{\itemsep}{0.5pt}
    \item We propose a human-annotation-free Automated Meta-cognition Evaluation framework, AutoMeco, benchmarking the meta-cognition lenses towards step rightness prediction.
    \item We propose a training-free Markovian Intrinsic Reward Adjustment strategy, MIRA, which enhances the LLM meta-cognition lenses by introducing stepwise dependencies.
    \item We conduct experiments on mathematical reasoning datasets with different difficulties (GSM8K, MATH500, and MinervaMATH), presenting the reasonableness of AutoMeco and the effectiveness of MIRA.
\end{itemize}

\section{Related Work}
\label{related_work}
\paragraph{Reasoning Step Error Detection Benchmarks} PRM800K~\cite{lightman2023let} classifies the intermediate reasoning step as positive, negative, or neutral, and underscores the importance of the reasoning step supervision to solve MATH problems. This urges the improvement of Process Reward Models, which are trained on datasets with annotations on step rightness, such as PRM800K, to output the step correctness probability.

Afterwards, MR-GSM8K~\cite{zeng2025mrgsmk} and MR-Math~\cite{xia2025evaluating} extend the step error detection beyond the classification task by manually annotating the reason behind the first error step based on subsets of GSM8K~\cite{cobbe2021training} and MATH~\cite{hendrycks2measuring} datasets, respectively. They propose the cruciality of the error reason interpretation ability of LLMs. Furthermore, BIG-Bench Mistake (BBM)~\cite{tyen2024llms} and MR-Ben~\cite{zeng2024mrben} extend the task to other domains instead of math problems only. BBM contains 2,186 instances from symbolic reasoning tasks~\cite{tyen2024llms}. MR-Ben consists of 5,975 instances covering natural sciences, coding, and logic~\cite{zeng2024mrben}. However, these benchmarks focus on the error detection ability of LLMs, which is a cognitive task. Instead, this paper aims to conduct meta-cognition evaluation on LLMs, delving into their intrinsic awareness of making mistakes during reasoning.

\paragraph{Process Reward Models} In the multi-step mathematical reasoning task, existing research mainly defines process reward models as classifiers that provide step correctness probability as fine-grained process supervision~\cite{uesato2022solving,lightman2023let,wang2024math,zhang2025lessons,shao2024deepseekmath}. \citet{lightman2023let} constructs PRM800k, a large-scale manually annotated process reward dataset, to train the PRM. To mitigate the dependence on costly human annotation, researchers propose automatic process supervision by annotating the step correctness using Monte Carlo Tree Search and training the process reward model on the automatically labeled datasets~\cite{wang2024math,luo2024improve,li2025enhancing}. Besides, \citet{li2025process} proposes the Process Q-Value Model (PQM) that considers the sequential dependencies between steps during the training of PRMs. Benefiting from the development of PRM, this paper utilizes it as a judge to annotate step correctness. Meanwhile, motivated by PQM, we propose a training-free reward adjustment strategy to enhance meta-cognition lenses.

\paragraph{LLM Self-Evaluation}
Self-evaluation in LLMs covers uncertainty estimation of LLMs. We focus on training-free measures of self-evaluation in this paper.
Traditional methods include token probability~\cite{jiang2021can}, perplexity~\cite{si2022prompting}, and entropy~\cite{huang2023look}. Furthermore, \citet{wang2025latent} proposes Chain-of-Embedding (CoE), which models the layer-by-layer changes of hidden states to reflect the answer correctness of LLMs and outperforms existing methods. However, existing self-evaluation measures mainly predict the answer correctness, and most methods ignore the connections among sentences or steps. Therefore, this paper concentrates on step correctness, regarding self-evaluation measures as the lenses of LLM meta-cognition, and endows their calculated intrinsic rewards with sequential dependencies.

Besides, sampling consistency is also proven effective in uncertainty quantification. \citet{manakul2023selfcheckgpt} calculates the similarity of multiple responses towards one question as the uncertainty score. \citet{tonolini2024bayesian} considers the response consistency towards multiple semantically equivalent queries.
However, we focus on methods based on LLM internal states in a single sample to study the intrinsic meta-cognition of LLMs.

\section{Methodology}
\label{method}

In this section, we first formally define the LLM meta-cognition observation task in \S\ref{subsec:task_definition}. Next, we present the Automated Meta-cognition Evaluation framework in \S\ref{subsec:AutoMeto}, which evaluates LLM meta-cognition lenses with the Process Reward Model as a judge. Finally, we introduce the Markovian Intrinsic Reward Adjustment strategy in \S\ref{subsec:reward_adjust} to enhance existing meta-cognition lenses.

\subsection{Task Definition}\label{subsec:task_definition}
The LLM meta-cognition observation task is defined as scoring the correctness of reasoning steps based on the internal states of LLMs. Given a question $Q$, an LLM generates a sequence containing a chain of reasoning steps $\{R_i\}_{i=1}^N$ and an answer $A$, where $N$ is the number of reasoning steps.
The LLM internal states of the reasoning step $R_i$ include hidden states $\boldsymbol{H}_i$, logits $\boldsymbol{Z}_i$, and probabilities $\boldsymbol{P}_i$ of the output tokens. The hidden states represent all-layer hidden states of each generated token, as shown in Equation~\ref{eq:hidden_state}.
\begin{equation}\label{eq:hidden_state}
    \boldsymbol{H}_i = \{\boldsymbol{h}_t\}_{t=1}^{T_i}, \quad\boldsymbol{h}_t = [\boldsymbol{h}_t^0,  ..., \boldsymbol{h}_t^{L}],
\end{equation}
where $T_i$ and $L$ denote the sequence length of $R_i$ and the number of LLM hidden layers, and $\boldsymbol{h}_t^0$ represents the embedding layer output of the $t$-th token. 
With the hidden states, the logits are the unnormalized output scores generated by the final hidden layer of the LLM in Equation~\ref{eq:logits}.
\begin{equation}\label{eq:logits}
    \boldsymbol{Z}_i=\{\boldsymbol{z}_t\}_{t=1}^{T_i}, \quad\boldsymbol{z}_t = \boldsymbol{h}_t^L \cdot \boldsymbol{W}_{\text{vocab}}^\top,
\end{equation}
where \( \boldsymbol{W}_{\text{vocab}} \in \mathbb{R}^{V \times d} \) is the vocabulary projection matrix, \( V \) is the vocabulary size, and \( d \) is the hidden dimension of the final layer. 
The probabilities of the output tokens are obtained by applying the softmax function to the logits:
\begin{equation}
\label{eq:probability}
    \boldsymbol{P}_i=\{\boldsymbol{p}_t\}_{t=1}^{T_i}, \quad\boldsymbol{p}_t = \textrm{softmax}(\boldsymbol{z}_t), \nonumber
\end{equation}
where the probability vector \( \boldsymbol{p}_t \in \mathbb{R}^V \) represents a normalized distribution over the vocabulary, guiding the final token prediction.

We formulate a self-evaluation method as a function $\mathcal{F}$ of the internal states, which calculates the confidence score $s_i$ of the reasoning steps $R_i$. 
\begin{equation}
    s_i = \mathcal{F}(\boldsymbol{H}_i, \boldsymbol{Z}_i, \boldsymbol{P}_i)
\end{equation}

\subsection{AutoMated Meta-Cognition Evaluation}\label{subsec:AutoMeto}

Our framework AutoMeco operates through four coordinated phases, as formalized in Algorithm~\ref{alg:framework}. The process initiates with structured response generation, followed by intrinsic rewarding with meta-cognition lenses, automated step correctness annotation using PRM, and metrics calculation.

\begin{algorithm}[ht]
\DontPrintSemicolon
\caption{Automated Meta-cognition Evaluation (AutoMeco)} \label{alg:framework}
\KwIn{Dataset $\mathcal{D}$, Language model $\mathcal{M}$, Threshold $\theta$}
\KwOut{Evaluation metrics}
\BlankLine
\For{each question $Q \in \mathcal{D}$}{
\tcc{\textbf{Structured Response Generation}}
$\boldsymbol{R} \leftarrow \mathcal{M}(Q)$\\
$\{R_i\}_{i=1}^N \leftarrow $ Segment $\boldsymbol{R}$ via boundary detection 

\tcc{\textbf{Stepwise State Aggregation}}
\For{each step $R_i \in \{R_1,...,R_N\}$}{
    Extract hidden states $\boldsymbol{H}_i = \{\{\boldsymbol{h}_t^k\}_{k=1}^L\}_{t=1}^{T_i}$ \\
    Collect logits $\boldsymbol{Z}_i$ and probabilities 
    $\boldsymbol{P}_i$
    
    \tcc{\textbf{Intrinsic Rewarding}}
    Compute confidence scores: $s_i^{\text{pred}} = \mathcal{F}(\boldsymbol{H}_i, \boldsymbol{Z}_i, \boldsymbol{P}_i)$
    }
\tcc{\textbf{Step Correctness Annotation}}
$\{s_i^{\text{true}}\}_{i=1}^N \leftarrow \mathcal{PRM}(Q, R_{1:N})$ \\
\For{$i=1$ \KwTo $N$}{
    \If{$s_i^{\text{true}} < \theta $}{$y_i^{\text{true}} \leftarrow 0$}
    \Else{$y_i^{\text{true} } \leftarrow 1$} } 
}
\tcc{\textbf{Metric Calculation}}
metrics $\leftarrow$ AUROC, AUPR, FPR95 \\
\Return metrics
\end{algorithm}

\paragraph{Structured Response Generation} Given an input question $Q$, the language model generates a response containing $N$ logically discrete reasoning steps $\{R_i\}_{i=1}^N$. Each step $R_i$ consists of consecutive tokens $[r_{1}, r_2, ..., r_{{T_i}}]$ representing a coherent reasoning unit. We employ boundary detection based on transitional phrases (e.g., ``Step 1:'', ``Step 2:'', and ``Answer:'') to segment the token sequence into interpretable steps.

\paragraph{Stepwise State Aggregation} For each identified step $R_i$, we aggregate the internal states across its constituent tokens. The hidden states of all the layers and every token are recorded:
\begin{equation}
    \boldsymbol{H}_i = \{ \{\boldsymbol{h}_t^k\}_{k=1}^L \}_{t=1}^{T_i}, \nonumber
\end{equation}
where $\boldsymbol{h}_t^k$ denotes the $k$-th layer's hidden state at token position $t$ in $R_i$. The step-level logits $\boldsymbol{z}_t$ and probabilities $\boldsymbol{p}_t$  are extracted from every token position within $R_i$, resulting in $\boldsymbol{Z}_i$ and $\boldsymbol{P}_i$.

\paragraph{Intrinsic Rewarding} The intrinsic reward $s_i^{\text{pred}}$ for each step is computed according to the specific meta-cognition lens $\mathcal{F}$:
\begin{equation}
    s_i^{\text{pred}} = \mathcal{F}(\boldsymbol{H}_i, \boldsymbol{Z}_i, \boldsymbol{P}_i) \nonumber
\end{equation}

\paragraph{Automated Step Correctness Annotation} The PRM receives the original question $Q$ and generated steps $\{R_i\}_{i=1}^N$ as input, calculating quality scores $\{s_i^{\text{true}}\}_{i=1}^N \in [0,1]^N$, as depicted in Equation~\ref{eq:prm-scoring}. These quality scores further produce binary labels based on a threshold $\theta$.
\begin{equation}\label{eq:prm-scoring}
    \{s_i^{\text{true}}\}_{i=1}^N = \mathcal{PRM}(Q,R_{1:N})
\end{equation}
\begin{equation}
    y_i^{\text{true}} = 
    \begin{cases}
        0,\ \text{if}\ s_i^{\text{true}}<\theta,\\
        1,\ \text{else, }
    \end{cases}
    \ \forall\ i \in \{1,...,N\}
\end{equation}
\paragraph{Metric Formalization} 
Following~\citet{wang2025latent}, we choose area under the precision-recall curve (AUPR)~\cite{manning1999foundations}, area under the receiver operating characteristic curve (AUROC)~\cite{boyd2013area}, and the false positive rate at 95\% true positive rate (FPR95)~\cite{wang2025latent} to evaluate the alignment between intrinsic confidence scores and process quality labels.
These metrics provide complementary views: AUROC measures global ranking consistency, FPR95 quantifies false alarm rates under high recall constraints, and AUPR evaluates precision-recall tradeoffs.

\subsection{Markovian Intrinsic Reward Adjustment }\label{subsec:reward_adjust}
The stepwise Markovian Intrinsic Reward Adjustment method models the reasoning as a Markov decision process and tunes the self-contained confidence scores by considering the sequential relations among the reasoning steps. We formalize the stepwise adjustment method in Algorithm~\ref{alg:reward_adj} and introduce it as follows.

\begin{algorithm}[h]
\DontPrintSemicolon
\caption{Markovian Intrinsic Reward Adjustment (MIRA)} \label{alg:reward_adj}
\KwIn{Reasoning trajectory $\tau=\{(\mathcal{S}_1,R_1),...,(\mathcal{S}_N,R_N),(\mathcal{S}_{N+1},A)\}$, 
    confidence scores $\{s_i^{\text{pred}}\}_{i=1}^N$, 
    discount factor $\gamma$}
\KwOut{Adjusted scores $\{\hat{s}_i^{\text{pred}}\}_{i=1}^N$}
\BlankLine

\tcc{\textbf{State Transition Modeling}}
\For{$i \leftarrow 1$ \KwTo $N$}{
    $\mathcal{S}_{i+1} \leftarrow \textrm{concat}(\mathcal{S}_i, R_i)$ 
}

\tcc{\textbf{Q-Value Backward Propagation}}
Initialize $V(\mathcal{S}_{N+1}) \leftarrow 0$\\
\For{$i \leftarrow N$ \KwTo $1$}{
    $\mathcal{Q}(\mathcal{S}_i,R_i) \leftarrow s_i^{\text{pred}} + \gamma \cdot V(\mathcal{S}_{i+1})$\\
    $V(\mathcal{S}_i) \leftarrow \max_{R_i} \mathcal{Q}(\mathcal{S}_i,R_i)$ 
}

\tcc{\textbf{Score Normalization}}
\For{$i \leftarrow 1$ \KwTo $N$}{
$\hat{s}_i^{\text{pred}} \leftarrow \frac{\exp(\mathcal{Q}(\mathcal{S}_i,R_i))}{\sum_{j=1}^N \exp(\mathcal{Q}(\mathcal{S}_j,R_j))}$
}
\Return $\{\hat{s}_i^{\text{pred}}\}_{i=1}^N$
\end{algorithm}

Given a reasoning trajectory $\tau$ defined as Equation~\ref{reasoning_traj} and~\ref{state-definition},  our training-free adjustment strategy formalizes the reasoning process as a deterministic MDP to adjust the step-level rewards by considering the interdependencies between steps.
\begin{equation}\label{reasoning_traj}
    \tau=\{(\mathcal{S}_1,R_1), ..., (\mathcal{S}_N, R_N), (\mathcal{S}_{N+1}, A) \}
\end{equation}
\begin{equation}\label{state-definition}
    \mathcal{S}_i=
    \begin{cases}
        Q, \ \text{if}\ \ i=1,\\
        \textrm{concat}(Q,R_1, ..., R_{i-1}), \ \text{else.}
    \end{cases}
\end{equation}
\paragraph{State Transition Modeling} The concatenation operation propagates state representations as shown in Equation~\ref{state_propogate}, where $\mathcal{S}_i$ denotes the $i$-th reasoning state and $R_{i+1}$ represents the corresponding reasoning action (i.e., the next reasoning step).
\begin{equation}\label{state_propogate}
    \mathcal{S}_{i+1} = \textrm{concat}(\mathcal{S}_i, R_{i}),\ \forall\ i \in \{1,...,N\}
\end{equation}

\paragraph{Q-Value Estimation and Backward Propagation} The expected future reward $\mathcal{Q}(\mathcal{S}_i,R_i)$ integrates immediate confidence and discounted future value in Equation~\ref{q-value-back}, with value function $V(\mathcal{S}_{i+1}) = \max_{R_{i+1}} \mathcal{Q}(\mathcal{S}_{i+1}, R_{i+1})$ and $\gamma \in (0,1]$ as the discount factor. In this equation, we recursively update Q-values from terminal state $\mathcal{S}_{N+1}$ with $V(\mathcal{S}_{N+1})=0$. This simplifies under a deterministic MDP to direct value assignment in Equation~\ref{q-value-back-simple}.

\begin{equation}\label{q-value-back}
    \mathcal{Q}(\mathcal{S}_i,R_i) = s_i^{\text{pred}} + \gamma \cdot \mathbb{E}_{\mathcal{S}_{i+1}}[V(\mathcal{S}_{i+1})]
\end{equation}
\begin{equation}\label{q-value-back-simple}
    \mathcal{Q}(\mathcal{S}_i,R_i) = s_i^{\text{pred}} + \gamma \cdot V(\mathcal{S}_{i+1})
\end{equation}

\paragraph{Score Normalization} Final adjusted scores are computed via softmax scaling:
\begin{equation}
    \hat{s}_i^{\text{pred}} = \frac{\exp(\mathcal{Q}(\mathcal{S}_i,R_i))}{\sum_{j=1}^N \exp(\mathcal{Q}(\mathcal{S}_i,R_i))} \nonumber
\end{equation}

\section{Experiment}
\label{experiment}
This section will answer four logically connected questions to present our experiments. \S\ref{exp:stat} explores whether it is statistically feasible to predict step correctness based on internal states. Furthermore, \S\ref{exp:comparison} answers two questions: (1) How do existing meta-cognition lenses perform towards the meta-cognition observation? (2) Is our proposed stepwise adjustment method effective in improving these meta-cognition lenses?  Afterwards, \S\ref{exp:prm} validates whether PRM-as-a-Judge is a reasonable method towards the meta-cognition evaluation.

\subsection{Experiments Setup}
\paragraph{Datasets} We focus on mathematical reasoning task with different difficulties. We evaluate the existing meta-cognition lenses on three datasets: the grade school math problems from GSM8K~\cite{cobbe2021training}, competition mathematics problems from MATH500~\cite{hendrycks2measuring}, and undergraduate- or Olympiad-level mathematical problems from MinervaMATH~\cite{lewkowycz2022solving}. We choose the first 250 problems of GSM8K due to its relatively low difficulty, which is also the English split of the multilingual math word problems from MGSM~\cite{shilanguage}. MATH500 contains 500 pieces selected by~\citet{lightman2023let} from the MATH dataset. MinervaMATH includes 272 problems that require quantitative reasoning.

\paragraph{Backbone Models} For PRM, we use Qwen2.5-Math-PRM-7B~\cite{zhang2025lessons} in our experiments, as it is one of the best mathematical process reward models according to recent research by ~\citet{zheng2024processbench}. For LLMs, we consider three open-sourced models, Qwen2.5-7B~\cite{qwen2}, Llama-3-8B-Instruct~\cite{llama3modelcard}, and Mistral-7B-Instruct~\cite{jiang2023mistral7b}, to conduct reasoning on the three datasets above. 

\paragraph{Baselines} We conduct six meta-cognition lenses to predict the step correctness, which results in the intrinsic rewards of each step in the reasoning chains. These baselines are: (1) CoE-C~\cite{wang2025latent}; (2) CoE-R~\cite{wang2025latent}; (3) $\Delta$Entropy~\cite{yin-etal-2024-reasoning}; (4) Max Probability (Maxprob); (5) Perplexity (PPL)~\cite{huang2023look}; (6) Entropy~\cite{si2022prompting}. Among them, (1) and (2) require access to hidden states of LLMs, and (3)-(6) only require probability distributions. More details are in Appendix~\ref{sec:self-evaluation-methods}.

\paragraph{Implementation Details} We set the maximal number of new tokens for Qwen2.5-7B as 768 and those for the other two models as 2048 because of their extra instruction-tuning. In the AutoMeco evaluation, we set $\rm{temperature}=1.0$ and $\rm{num\_sequences}=1$ to evaluate the model under the condition of greedy generation. In the Best-of-N (BoN) evaluation, we set $\rm{temperature}=0.8$ and $\rm{N}=6$. We select the best response from the $N$ samples by choosing the one with the highest intrinsic reward averaged over steps. For Mistral-7B-Instruct, we use Mistral-7B-Instruct-v0.2~\cite{mistral}. Appendix~\ref{sec:prompt} shows the prompt templates for the three datasets. All the experiments are conducted on four 24G 3090 GPUs or one A100 80G GPU. For clear visualization, we use the kdeplot function from the seaborn libary~\cite{waskom2021seaborn} with contour levels of 7, density thresholds of 0.15 for correct and 0.1 for incorrect samples, and $\textrm{bw\_adjust}=1.5$.

\subsection{Statistical Feasibility}
\label{exp:stat}
We prompt Qwen2.5-7B to reason on GSM8K, MATH500, and MinervaMATH. We conduct the self-evaluation measures to predict the step correctness, which results in the intrinsic rewards of each step in the reasoning chains. Subsequently, with the step rewards annotated by PRM, we calculate the correlation of the intrinsic and PRM rewards on the data split conditioned on PRM reward$\in (0,1]\cup [0.9, 1)$. PRM denotes Qwen2.5-PRM-Math-7B in the following, if without reclarification. Besides, we show the distinguishability of step correctness by visualizing the kernel density estimation of Magnitude and Angle, two internal features proposed by~\citet{wang2025latent}. Appendix~\ref{sec:self-evaluation-methods} presents details of the two features.

\begin{figure}[H]
\centering
\begin{subfigure}[t]{0.32\textwidth}
    \centering
    \includegraphics[width=1\linewidth]{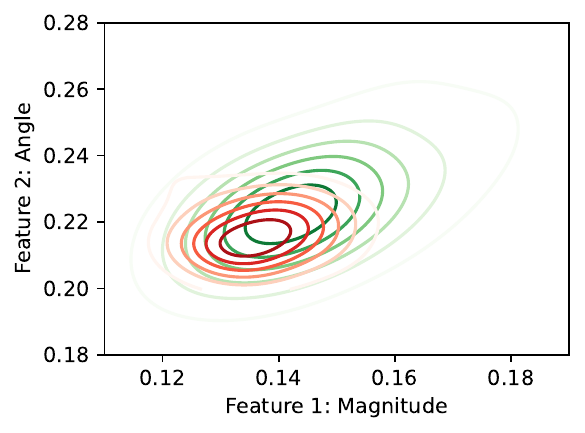}
    \caption{GSM8K}
    \label{fig:dataset1}
\end{subfigure}
\hfill
\begin{subfigure}[t]{0.32\textwidth}
    \centering
    \includegraphics[width=1\linewidth]{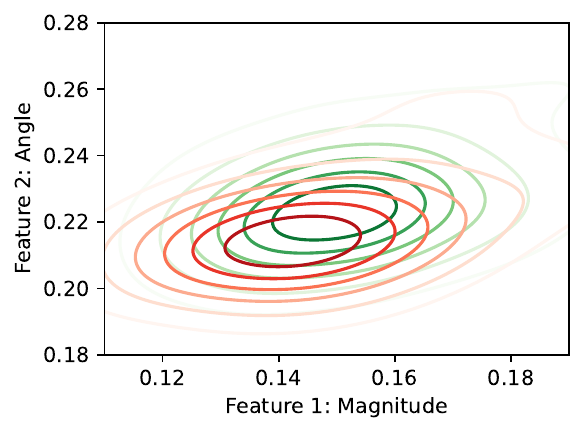}
    \caption{MATH500}
    \label{fig:dataset2}
\end{subfigure}
\hfill
\begin{subfigure}[t]{0.32\textwidth}
    \centering
    \includegraphics[width=1\linewidth]{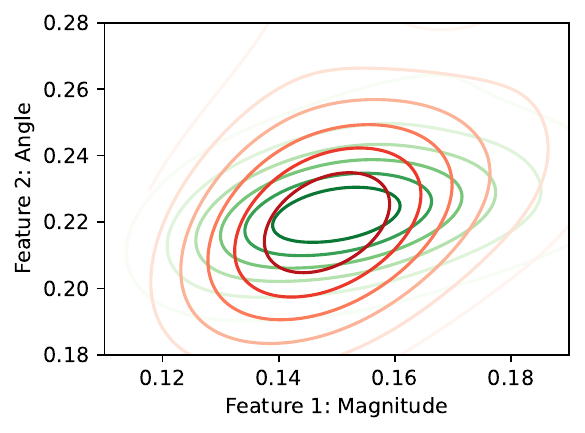}
    \caption{MinervaMATH}
    \label{fig:dataset3}
\end{subfigure}
\caption{Intrinsic feature distributions of correct and incorrect steps of Qwen2.5-7B on GSM8K, MATH500, and MinervaMATH. \green{Green} and \red{red} contours represent features of \green{correct} and \red{wrong} steps.}
\label{fig:visual}
\end{figure}

Table~\ref{tab:correlation} presents that the intrinsic and PRM rewards have a significant positive correlation on the GSM8K dataset, which proves the feasibility of LLM meta-cognition observation. However, the premise is that the LLM can handle the task to some extent, as the correlation drops significantly with the rise of task difficulty. Furthermore, entropy is statistically the most promising method to capture the inherent presentation of step correctness, which consistently has the highest correlation coefficients on the three datasets. Besides, Figure~\ref{fig:visual} illustrates the feature distributions of the correct and wrong steps. It shows the decline of the step correctness predictability when the dataset gets harder, which is consistent with the statistics above.
\begin{table}[htbp]
\centering
\caption{Spearman coefficient~\cite{spearman1961proof} and Kendall's Tau~\cite{kendall1938new} between the intrinsic and PRM rewards of Qwen-2.5-7B on three datasets. All values are formatted as \texttt{coefficient (p-value)} with p-values being in scientific notation, retaining two significant figures (e.g., 1.7e-29 denotes $1.7 \times 10^{-29}$). \textit{S} and \textit{K} stands for Spearman and Kendall, respectively. Bold and underlined denote the best and second-best.}
\renewcommand\arraystretch{1.85}
\label{tab:correlation}
\resizebox{\linewidth}{!}{
\begin{tabular}{ccccccc}

\toprule
\multirow{2}{*}{\textbf{Methods}} & \multicolumn{2}{c}{\textbf{GSM8K}} & \multicolumn{2}{c}{\textbf{MATH500}} & \multicolumn{2}{c}{\textbf{MinervaMATH}} \\
\cmidrule(lr){2-3} \cmidrule(lr){4-5} \cmidrule(lr){6-7}
  & \textit{S} & \textit{K} & \textit{S} & \textit{K} & \textit{S} & \textit{K} \\
\hline
\texttt{CoE-C} & \makecell[c]{0.040 \\ {\footnotesize (0.174)}} 
& \makecell[c]{0.035 \\ {\footnotesize (0.074)}} 
& \makecell[c]{0.068 \\ {\footnotesize (0.0001)}} 
& \makecell[c]{0.046 \\ {\footnotesize (0.0001)}} 
& \makecell[c]{0.010 \\ {\footnotesize (0.710)}} 
& \makecell[c]{0.007 \\ {\footnotesize (0.685)}} \\
\hline
\texttt{CoE-R} & \makecell[c]{0.325 \\ {\footnotesize (1.7e-29)}} 
& \makecell[c]{0.225 \\ {\footnotesize (4.1e-30)}} 
& \makecell[c]{0.212 \\ {\footnotesize (1.4e-33)}} 
& \makecell[c]{0.145 \\ {\footnotesize (2.2e-34)}} 
& \makecell[c]{0.255 \\ {\footnotesize (3.7e-21)}} 
& \makecell[c]{0.175 \\ {\footnotesize (1.7e-21)}} \\
\hline
\texttt{Maxprob} & \makecell[c]{0.498 \\ {\footnotesize (1.3e-72)}} 
& \makecell[c]{0.354 \\ {\footnotesize (7.3e-72)}} 
& \makecell[c]{0.241 \\ {\footnotesize (2.1e-43)}} 
& \makecell[c]{0.164 \\ {\footnotesize (8.1e-44)}} 
& \makecell[c]{\underline{0.265} \\ {\footnotesize (9.6e-23)}} 
& \makecell[c]{\underline{0.182} \\ {\footnotesize (3.1e-23)}} \\
\hline
\texttt{PPL} & \makecell[c]{\underline{0.499} \\ {\footnotesize (6.1e-73)}} 
& \makecell[c]{\underline{0.355} \\ {\footnotesize (4.9e-72)}} 
& \makecell[c]{\underline{0.243} \\ {\footnotesize (4.3e-44)}} 
& \makecell[c]{\underline{0.165} \\ {\footnotesize (2.6e-44)}} 
& \makecell[c]{0.263 \\ {\footnotesize (2.2e-22)}} 
& \makecell[c]{0.180 \\ {\footnotesize (9.1e-23)}} \\
\hline
\texttt{Entropy} & \makecell[c]{\textbf{0.521} \\ {\footnotesize (1.6e-80)}} 
& \makecell[c]{\textbf{0.370} \\ {\footnotesize (3.1e-78)}} 
& \makecell[c]{\textbf{0.246} \\ {\footnotesize (3.0e-45)}} 
& \makecell[c]{\textbf{0.168} \\ {\footnotesize (1.1e-45)}} 
& \makecell[c]{\textbf{0.270} \\ {\footnotesize (1.7e-23)}} 
& \makecell[c]{\textbf{0.185} \\ {\footnotesize (6.8e-24)}} \\
\hline
\texttt{$\Delta$Entropy} & \makecell[c]{ -0.060  \\ {\footnotesize (0.041)}} 
& \makecell[c]{-0.039 \\ {\footnotesize (0.052)}} 
& \makecell[c]{0.147 \\ {\footnotesize (7.5e-17)}} 
& \makecell[c]{0.097 \\ {\footnotesize (1.7e-16)}} 
& \makecell[c]{0.066 \\ {\footnotesize (0.017)}} 
& \makecell[c]{0.044 \\ {\footnotesize  (0.016)}} \\
\bottomrule
\end{tabular}
}
\end{table}

\subsection{Comparison and Ablation Study}
\label{exp:comparison}
We evaluate the meta-cognition lenses in a more challenging and valuable setting with the three language models. In this setting, we test the ability of these methods to distinguish the correct and incorrect steps, including wrong and unsure ones, instead of merely the wrong steps.

\begin{table*}[h]
\centering
\caption{AutoMeco and Best-of-N evaluation results of meta-cognition lenses with and without our proposed reward adjustment strategy, MIRA, across three mathematics reasoning datasets and three LLMs. \green{Green} and \red{red} denote whether MIRA improves the meta-cognition lenses. ``Acc'' denotes accuracy. ``Majority'' represents majority voting. ``PRM'' denotes voting based on the step-averaged rewards calculated by the PRM. }
\label{fig:comparison}
\renewcommand\arraystretch{1.3}

\resizebox{\linewidth}{!}{
\begin{tabular}{c|cccc|cccc|cccc}
    \bottomrule
    \multirow{3}[4]{*}{\textbf{Methods}} & \multicolumn{4}{c|}{\textbf{Qwen2.5-7B}} & \multicolumn{4}{c|}{\textbf{Llama-3-8B-Instruct}} & \multicolumn{4}{c}{\textbf{Mistral-7B-Instruct}} \\
\cline{2-13}          & \multirow{2}[2]{*}{Best-of-N Acc$\uparrow$} & \multirow{2}[2]{*}{AUROC$\uparrow$} & \multirow{2}[2]{*}{FPR95$\downarrow$} & \multirow{2}[2]{*}{AUPR$\uparrow$} & \multirow{2}[2]{*}{Best-of-N Acc$\uparrow$} & \multirow{2}[2]{*}{AUROC$\uparrow$} & \multirow{2}[2]{*}{FPR95$\downarrow$} & \multirow{2}[2]{*}{AUPR$\uparrow$} & \multirow{2}[2]{*}{Best-of-N Acc$\uparrow$} & \multirow{2}[2]{*}{AUROC$\uparrow$} & \multirow{2}[2]{*}{FPR95$\downarrow$} & \multirow{2}[2]{*}{AUPR$\uparrow$} \\
          &       &       &       &       &       &       &       &       &       &       &       &  \\
    \hline
    \rowcolor{gray!40}\multicolumn{13}{c}{GSM8K} \\
    \hline
    \multicolumn{1}{l|}{Maxprob} & \multicolumn{1}{l}{53.60 } & \multicolumn{1}{l}{61.25 } & \multicolumn{1}{l}{90.74 } & \multicolumn{1}{l|}{96.82 } & \multicolumn{1}{l}{73.20 } & \multicolumn{1}{l}{65.53 } & \multicolumn{1}{l}{90.29 } & \multicolumn{1}{l|}{94.67 } & \multicolumn{1}{l}{42.80 } & \multicolumn{1}{l}{68.46 } & \multicolumn{1}{l}{86.17 } & \multicolumn{1}{l}{51.79 } \\
    \multicolumn{1}{l|}{+ MIRA (ours)} & \multicolumn{1}{l}{58.80 \textcolor{improve}{(+5.20)}} & \multicolumn{1}{l}{67.50 \textcolor{improve}{(+6.25)}} & \multicolumn{1}{l}{81.48 \textcolor{improve}{(-9.26)}} & \multicolumn{1}{l|}{96.88 \textcolor{improve}{(+0.06)}} & \multicolumn{1}{l}{73.60 \textcolor{improve}{(+0.40)}} & \multicolumn{1}{l}{58.21 \textcolor{degrade}{(-7.32)}} & \multicolumn{1}{l|}{85.44 \textcolor{improve}{(-4.85)}} & \multicolumn{1}{l|}{92.60 \textcolor{degrade}{(-2.07)}} & \multicolumn{1}{l}{45.20 \textcolor{improve}{(+2.40)}} & \multicolumn{1}{l}{58.88 \textcolor{degrade}{(-9.58)}} & \multicolumn{1}{l}{79.86 \textcolor{improve}{(-6.31)}} & \multicolumn{1}{l}{39.94 \textcolor{degrade}{(-11.85)}} \\
    \hline
    \multicolumn{1}{l|}{PPL} & \multicolumn{1}{l}{57.20 } & \multicolumn{1}{l}{61.19 } & \multicolumn{1}{l}{90.74 } & \multicolumn{1}{l|}{96.81 } & \multicolumn{1}{l}{80.40 } & \multicolumn{1}{l}{65.67 } & \multicolumn{1}{l}{90.29 } & \multicolumn{1}{l|}{94.70 } & \multicolumn{1}{l}{46.40 } & \multicolumn{1}{l}{68.93 } & \multicolumn{1}{l}{83.79 } & \multicolumn{1}{l}{51.68 } \\
    \multicolumn{1}{l|}{+ MIRA (ours)} & \multicolumn{1}{l}{66.40 \textcolor{improve}{(+9.20)}} & \multicolumn{1}{l}{70.92 \textcolor{improve}{(+9.73)}} & \multicolumn{1}{l}{79.63 \textcolor{improve}{(-11.11)}} & \multicolumn{1}{l|}{97.65 \textcolor{improve}{(+0.84)}} & \multicolumn{1}{l}{78.80 \textcolor{degrade}{(-1.60)}} & \multicolumn{1}{l}{59.32 \textcolor{degrade}{(-6.35)}} & \multicolumn{1}{l}{84.47 \textcolor{improve}{(-5.82)}} & \multicolumn{1}{l|}{94.60 \textcolor{degrade}{(-0.10)}} & \multicolumn{1}{l}{47.60 \textcolor{improve}{(+1.20)}} & \multicolumn{1}{l}{61.74 \textcolor{degrade}{(-7.19)}} & \multicolumn{1}{l}{79.02 \textcolor{improve}{(-4.77)}} & \multicolumn{1}{l}{64.54 \textcolor{improve}{(+12.86)}} \\
    \hline
    \multicolumn{1}{l|}{Entropy} & \multicolumn{1}{l}{56.40 } & \multicolumn{1}{l}{60.62 } & \multicolumn{1}{l}{92.59 } & \multicolumn{1}{l|}{96.84 } & \multicolumn{1}{l}{80.00 } & \multicolumn{1}{l}{66.99 } & \multicolumn{1}{l}{85.44 } & \multicolumn{1}{l|}{94.86 } & \multicolumn{1}{l}{44.80 } & \multicolumn{1}{l}{71.68 } & \multicolumn{1}{l}{78.43 } & \multicolumn{1}{l}{54.57 } \\
    \multicolumn{1}{l|}{+ MIRA (ours)} & \multicolumn{1}{l}{63.20 \textcolor{improve}{(+6.80)}} & \multicolumn{1}{l}{71.90 \textcolor{improve}{(+11.28)}} & \multicolumn{1}{l}{75.93 \textcolor{improve}{(-16.66)}} & \multicolumn{1}{l|}{97.56 \textcolor{improve}{(+0.72)}} & \multicolumn{1}{l}{79.20 \textcolor{degrade}{(-0.80)}} & \multicolumn{1}{l}{60.87 \textcolor{degrade}{(-6.12)}} & \multicolumn{1}{l}{86.41 \textcolor{degrade}{(+0.97)}} & \multicolumn{1}{l|}{93.67 \textcolor{degrade}{(-1.19)}} & \multicolumn{1}{l}{47.20 \textcolor{improve}{(+2.40)}} & \multicolumn{1}{l}{64.14 \textcolor{degrade}{(-7.54)}} & \multicolumn{1}{l}{77.00 \textcolor{improve}{(-1.43)}} & \multicolumn{1}{l}{54.58 \textcolor{improve}{(+0.01)}} \\
    \hline
    \multicolumn{1}{l|}{$\Delta$Entropy} & \multicolumn{1}{l}{60.00 } & \multicolumn{1}{l}{64.02 } & \multicolumn{1}{l}{94.44 } & \multicolumn{1}{l|}{96.82 } & \multicolumn{1}{l}{77.20 } & \multicolumn{1}{l}{50.45 } & \multicolumn{1}{l}{97.09 } & \multicolumn{1}{l|}{90.18 } & \multicolumn{1}{l}{43.60 } & \multicolumn{1}{l}{56.86 } & \multicolumn{1}{l}{83.67 } & \multicolumn{1}{l}{38.15 } \\
    \multicolumn{1}{l|}{+ MIRA (ours)} & \multicolumn{1}{l}{61.60 \textcolor{improve}{(+1.60)}} & \multicolumn{1}{l}{56.65 \textcolor{degrade}{(-7.57)}} & \multicolumn{1}{l}{94.44 \textcolor{improve}{(-0.00)}} & \multicolumn{1}{l|}{95.69 \textcolor{degrade}{(-1.13)}} & \multicolumn{1}{l}{74.40 \textcolor{degrade}{(-2.80)}} & \multicolumn{1}{l}{45.66 \textcolor{degrade}{(-4.79)}} & \multicolumn{1}{l}{98.28 \textcolor{degrade}{(+1.19)}} & \multicolumn{1}{l|}{90.19 \textcolor{improve}{(+0.01)}} & \multicolumn{1}{l}{49.60 \textcolor{improve}{(+6.00)}} & \multicolumn{1}{l}{55.77 \textcolor{degrade}{(-1.09)}} & \multicolumn{1}{l}{86.77 \textcolor{degrade}{(+3.10)}} & \multicolumn{1}{l}{40.23 \textcolor{improve}{(+2.08)}} \\
    \hline
    \multicolumn{1}{l|}{CoE-R} & \multicolumn{1}{l}{54.80 } & \multicolumn{1}{l}{64.78 } & \multicolumn{1}{l}{88.89 } & \multicolumn{1}{l|}{97.14 } & \multicolumn{1}{l}{77.60 } & \multicolumn{1}{l}{63.84 } & \multicolumn{1}{l}{91.26 } & \multicolumn{1}{l|}{94.32 } & \multicolumn{1}{l}{44.80 } & \multicolumn{1}{l}{39.24 } & \multicolumn{1}{l}{96.78 } & \multicolumn{1}{l}{32.47 } \\
    \multicolumn{1}{l|}{+ MIRA (ours)} & \multicolumn{1}{l}{75.60 \textcolor{improve}{(+20.80)}} & \multicolumn{1}{l}{65.23 \textcolor{improve}{(+0.45)}} & \multicolumn{1}{l}{88.89 \textcolor{improve}{(-0.00)}} & \multicolumn{1}{l|}{96.90 \textcolor{degrade}{(-0.24)}} & \multicolumn{1}{l}{77.60 \textcolor{improve}{(+0.00)}} & \multicolumn{1}{l}{63.52 \textcolor{degrade}{(-0.32)}} & \multicolumn{1}{l}{84.47 \textcolor{improve}{(-6.79)}} & \multicolumn{1}{l|}{94.40 \textcolor{improve}{(+0.08)}} & \multicolumn{1}{l}{39.60 \textcolor{degrade}{(-5.20)}} & \multicolumn{1}{l}{53.58 \textcolor{improve}{(+14.34)}} & \multicolumn{1}{l}{93.86 \textcolor{improve}{(-2.92)}} & \multicolumn{1}{l}{37.08 \textcolor{improve}{(+4.61)}} \\
    \hline
    \multicolumn{1}{l|}{CoE-C} & \multicolumn{1}{l}{58.80 } & \multicolumn{1}{l}{69.53 } & \multicolumn{1}{l}{90.74 } & \multicolumn{1}{l|}{97.65 } & \multicolumn{1}{l}{70.80 } & \multicolumn{1}{l}{72.29 } & \multicolumn{1}{l}{79.61 } & \multicolumn{1}{l|}{95.91 } & \multicolumn{1}{l}{44.00 } & \multicolumn{1}{l}{52.33 } & \multicolumn{1}{l}{93.56 } & \multicolumn{1}{l}{39.99 } \\
    \multicolumn{1}{l|}{+ MIRA (ours)} & \multicolumn{1}{l}{57.60 \textcolor{degrade}{(-1.20)}} & \multicolumn{1}{l}{68.87 \textcolor{degrade}{(-0.66)}} & \multicolumn{1}{l}{70.37 \textcolor{improve}{(-20.37)}} & \multicolumn{1}{l|}{97.05 \textcolor{degrade}{(-0.60)}} & \multicolumn{1}{l}{72.80 \textcolor{improve}{(+2.00)}} & \multicolumn{1}{l}{58.24 \textcolor{degrade}{(-14.05)}} & \multicolumn{1}{l}{88.35 \textcolor{degrade}{(+8.74)}} & \multicolumn{1}{l|}{92.71 \textcolor{degrade}{(-3.20)}} & \multicolumn{1}{l}{44.40 \textcolor{improve}{(+0.40)}} & \multicolumn{1}{l}{58.56 \textcolor{improve}{(+6.23)}} & \multicolumn{1}{l}{80.33 \textcolor{improve}{(-13.23)}} & \multicolumn{1}{l}{39.82 \textcolor{degrade}{(-0.17)}} \\
    \hline
    Majority\ /\ PRM & \multicolumn{1}{l}{86.80\ /\ 75.20} & -     & -     & -     & \multicolumn{1}{l}{86.80\ /\ 89.20}  & -     & -     & -     & \multicolumn{1}{l}{60.40\ /\ 70.80} & -     & -     & - \\
    \hline
    \rowcolor{gray!40}\multicolumn{13}{c}{MATH500} \\
    \hline
    \multicolumn{1}{l|}{Maxprob} & \multicolumn{1}{l}{35.00 } & \multicolumn{1}{l}{58.12 } & \multicolumn{1}{l}{95.92 } & \multicolumn{1}{l|}{91.43 } & \multicolumn{1}{l}{18.60 } & \multicolumn{1}{l}{63.76 } & \multicolumn{1}{l}{88.72 } & \multicolumn{1}{l|}{72.51 } & \multicolumn{1}{l}{6.20 } & \multicolumn{1}{l}{56.00 } & \multicolumn{1}{l}{94.72 } & \multicolumn{1}{l}{12.40 } \\
    \multicolumn{1}{l|}{+ MIRA (ours)} & \multicolumn{1}{l}{37.60 \textcolor{improve}{(+2.60)}} & \multicolumn{1}{l}{64.66 \textcolor{improve}{(+6.54)}} & \multicolumn{1}{l}{86.52 \textcolor{improve}{(-9.40)}} & \multicolumn{1}{l|}{92.52 \textcolor{improve}{(+1.09)}} & \multicolumn{1}{l}{18.00 \textcolor{degrade}{(-0.60)}} & \multicolumn{1}{l}{62.03 \textcolor{degrade}{(-1.73)}} & \multicolumn{1}{l}{87.26 \textcolor{improve}{(-1.46)}} & \multicolumn{1}{l|}{68.04 \textcolor{degrade}{(-4.47)}} & \multicolumn{1}{l}{5.60 \textcolor{degrade}{(-0.60)}} & \multicolumn{1}{l}{56.89 \textcolor{improve}{(+0.89)}} & \multicolumn{1}{l}{91.86 \textcolor{improve}{(-2.86)}} & \multicolumn{1}{l}{12.71 \textcolor{improve}{(+0.31)}} \\
    \hline
    \multicolumn{1}{l|}{PPL} & \multicolumn{1}{l}{40.00 } & \multicolumn{1}{l}{57.83 } & \multicolumn{1}{l}{95.61 } & \multicolumn{1}{l|}{91.43 } & \multicolumn{1}{l}{23.20 } & \multicolumn{1}{l}{63.84 } & \multicolumn{1}{l}{88.04 } & \multicolumn{1}{l|}{72.71 } & \multicolumn{1}{l}{6.00 } & \multicolumn{1}{l}{56.35 } & \multicolumn{1}{l}{94.07 } & \multicolumn{1}{l}{12.60 } \\
    \multicolumn{1}{l|}{+ MIRA (ours)} & \multicolumn{1}{l}{44.20 \textcolor{improve}{(+4.20)}} & \multicolumn{1}{l}{64.40 \textcolor{improve}{(+6.57)}} & \multicolumn{1}{l}{86.52 \textcolor{improve}{(-9.09)}} & \multicolumn{1}{l|}{93.88 \textcolor{improve}{(+2.45)}} & \multicolumn{1}{l}{21.00 \textcolor{degrade}{(-2.20)}} & \multicolumn{1}{l}{62.77 \textcolor{degrade}{(-1.07)}} & \multicolumn{1}{l}{86.48 \textcolor{improve}{(-1.56)}} & \multicolumn{1}{l|}{75.74 \textcolor{improve}{(+3.03)}} & \multicolumn{1}{l}{6.20 \textcolor{improve}{(+0.20)}} & \multicolumn{1}{l}{54.51 \textcolor{degrade}{(-1.84)}} & \multicolumn{1}{l}{91.70 \textcolor{improve}{(-2.37)}} & \multicolumn{1}{l}{41.29 \textcolor{improve}{(+28.69)}} \\
    \hline
    \multicolumn{1}{l|}{Entropy} & \multicolumn{1}{l}{40.20 } & \multicolumn{1}{l}{57.53 } & \multicolumn{1}{l}{96.65 } & \multicolumn{1}{l|}{91.38 } & \multicolumn{1}{l}{22.40 } & \multicolumn{1}{l}{64.04 } & \multicolumn{1}{l}{87.71 } & \multicolumn{1}{l|}{72.63 } & \multicolumn{1}{l}{6.20 } & \multicolumn{1}{l}{55.43 } & \multicolumn{1}{l}{92.97 } & \multicolumn{1}{l}{12.41 } \\
    \multicolumn{1}{l|}{+ MIRA (ours)} & \multicolumn{1}{l}{43.00 \textcolor{improve}{(+2.80)}} & \multicolumn{1}{l}{66.23 \textcolor{improve}{(+8.70)}} & \multicolumn{1}{l}{85.58 \textcolor{improve}{(-11.07)}} & \multicolumn{1}{l|}{93.19 \textcolor{improve}{(+1.81)}} & \multicolumn{1}{l}{19.60 \textcolor{degrade}{(-2.80)}} & \multicolumn{1}{l}{65.87 \textcolor{improve}{(+1.83)}} & \multicolumn{1}{l}{86.48 \textcolor{improve}{(-1.23)}} & \multicolumn{1}{l|}{74.98 \textcolor{improve}{(+2.35)}} & \multicolumn{1}{l}{5.60 \textcolor{degrade}{(-0.60)}} & \multicolumn{1}{l}{53.20 \textcolor{degrade}{(-2.23)}} & \multicolumn{1}{l}{92.88 \textcolor{improve}{(-0.09)}} & \multicolumn{1}{l}{23.52 \textcolor{improve}{(+11.11)}} \\
    \hline
    \multicolumn{1}{l|}{$\Delta$Entropy} & \multicolumn{1}{l}{41.40 } & \multicolumn{1}{l}{54.21 } & \multicolumn{1}{l}{97.81 } & \multicolumn{1}{l|}{90.91 } & \multicolumn{1}{l}{17.80 } & \multicolumn{1}{l}{53.77 } & \multicolumn{1}{l}{96.42 } & \multicolumn{1}{l|}{62.51 } & \multicolumn{1}{l}{4.80 } & \multicolumn{1}{l}{49.63 } & \multicolumn{1}{l}{95.54 } & \multicolumn{1}{l}{12.60 } \\
    \multicolumn{1}{l|}{+ MIRA (ours)} & \multicolumn{1}{l}{37.20 \textcolor{degrade}{(-4.20)}} & \multicolumn{1}{l}{64.62 \textcolor{improve}{(+10.41)}} & \multicolumn{1}{l}{89.03 \textcolor{improve}{(-8.78)}} & \multicolumn{1}{l|}{92.69 \textcolor{improve}{(+1.78)}} & \multicolumn{1}{l}{18.00 \textcolor{improve}{(+0.20)}} & \multicolumn{1}{l}{62.70 \textcolor{improve}{(+8.93)}} & \multicolumn{1}{l}{90.61 \textcolor{improve}{(-5.81)}} & \multicolumn{1}{l|}{68.37 \textcolor{improve}{(+5.86)}} & \multicolumn{1}{l}{7.20 \textcolor{improve}{(+2.40)}} & \multicolumn{1}{l}{61.45 \textcolor{improve}{(+11.82)}} & \multicolumn{1}{l}{90.76 \textcolor{improve}{(-4.78)}} & \multicolumn{1}{l}{17.51 \textcolor{improve}{(+4.91)}} \\
    \hline
    \multicolumn{1}{l|}{CoE-R} & \multicolumn{1}{l}{39.00 } & \multicolumn{1}{l}{47.82 } & \multicolumn{1}{l}{97.81 } & \multicolumn{1}{l|}{88.17 } & \multicolumn{1}{l}{19.80 } & \multicolumn{1}{l}{52.37 } & \multicolumn{1}{l}{92.07 } & \multicolumn{1}{l|}{60.95 } & \multicolumn{1}{l}{6.00 } & \multicolumn{1}{l}{45.37 } & \multicolumn{1}{l}{94.99 } & \multicolumn{1}{l}{12.99 } \\
    \multicolumn{1}{l|}{+ MIRA (ours)} & \multicolumn{1}{l}{44.00 \textcolor{improve}{(+5.00)}} & \multicolumn{1}{l}{60.43 \textcolor{improve}{(+12.61)}} & \multicolumn{1}{l}{85.89 \textcolor{improve}{(-11.92)}} & \multicolumn{1}{l|}{91.64 \textcolor{improve}{(+2.47)}} & \multicolumn{1}{l}{17.60 \textcolor{degrade}{(-2.20)}} & \multicolumn{1}{l}{57.96 \textcolor{improve}{(+5.59)}} & \multicolumn{1}{l}{92.07 \textcolor{improve}{(-0.00)}} & \multicolumn{1}{l|}{64.62 \textcolor{improve}{(+3.67)}} & \multicolumn{1}{l}{4.40 \textcolor{degrade}{(-1.60)}} & \multicolumn{1}{l}{50.72 \textcolor{improve}{(+5.35)}} & \multicolumn{1}{l}{93.15 \textcolor{improve}{(-1.84)}} & \multicolumn{1}{l}{11.27 \textcolor{degrade}{(-1.72)}} \\
    \hline
    \multicolumn{1}{l|}{CoE-C} & \multicolumn{1}{l}{37.40 } & \multicolumn{1}{l}{59.71 } & \multicolumn{1}{l}{94.36 } & \multicolumn{1}{l|}{91.35 } & \multicolumn{1}{l}{16.80 } & \multicolumn{1}{l}{59.27 } & \multicolumn{1}{l}{93.52 } & \multicolumn{1}{l|}{68.83 } & \multicolumn{1}{l}{6.60 } & \multicolumn{1}{l}{59.80 } & \multicolumn{1}{l}{86.22 } & \multicolumn{1}{l}{13.43 } \\
    \multicolumn{1}{l|}{+ MIRA (ours)} & \multicolumn{1}{l}{37.40 \textcolor{improve}{(+0.00)}} & \multicolumn{1}{l}{65.03 \textcolor{improve}{(+5.32)}} & \multicolumn{1}{l}{84.64 \textcolor{improve}{(-9.28)}} & \multicolumn{1}{l|}{92.69 \textcolor{improve}{(+1.34)}} & \multicolumn{1}{l}{17.80 \textcolor{improve}{(+1.00)}} & \multicolumn{1}{l}{61.32 \textcolor{improve}{(+2.05)}} & \multicolumn{1}{l}{89.39 \textcolor{improve}{(-4.13)}} & \multicolumn{1}{l|}{67.65 \textcolor{degrade}{(-1.18)}} & \multicolumn{1}{l}{6.00 \textcolor{degrade}{(-0.60)}} & \multicolumn{1}{l}{57.75 \textcolor{degrade}{(-2.05)}} & \multicolumn{1}{l}{90.76 \textcolor{degrade}{(+4.54)}} & \multicolumn{1}{l}{13.03 \textcolor{degrade}{(-0.40)}} \\
    \hline
    Majority\ /\ PRM & \multicolumn{1}{l}{51.80\ /\ 49.60}  & -     & -     & -     & \multicolumn{1}{l}{24.60\ /\ 31.20}   & -     & -     & -      & \multicolumn{1}{l}{9.20\ /\ 11.80}  & -     & -     & -     \\
    \hline
    \rowcolor{gray!40}\multicolumn{13}{c}{MinervaMATH} \\
    \hline
    \multicolumn{1}{l|}{Maxprob} & \multicolumn{1}{l}{8.46 } & \multicolumn{1}{l}{52.75 } & \multicolumn{1}{l}{97.01 } & \multicolumn{1}{l|}{88.23 } & \multicolumn{1}{l}{6.62 } & \multicolumn{1}{l}{56.13 } & \multicolumn{1}{l}{92.51 } & \multicolumn{1}{l|}{72.04 } & \multicolumn{1}{l}{1.47 } & \multicolumn{1}{l}{53.33 } & \multicolumn{1}{l}{91.91 } & \multicolumn{1}{l}{17.24 } \\
    \multicolumn{1}{l|}{+ MIRA (ours)} & \multicolumn{1}{l}{9.56 \textcolor{improve}{(+1.10)}} & \multicolumn{1}{l}{63.37 \textcolor{improve}{(+10.62)}} & \multicolumn{1}{l}{94.61 \textcolor{improve}{(-2.40)}} & \multicolumn{1}{l|}{90.57 \textcolor{improve}{(+2.34)}} & \multicolumn{1}{l}{6.25 \textcolor{degrade}{(-0.37)}} & \multicolumn{1}{l}{60.18 \textcolor{improve}{(+4.05)}} & \multicolumn{1}{l}{92.90 \textcolor{degrade}{(+0.39)}} & \multicolumn{1}{l|}{71.33 \textcolor{degrade}{(-0.71)}} & \multicolumn{1}{l}{1.47 \textcolor{improve}{(+0.00)}} & \multicolumn{1}{l}{60.20 \textcolor{improve}{(+6.87)}} & \multicolumn{1}{l}{88.14 \textcolor{improve}{(-3.77)}} & \multicolumn{1}{l}{18.82 \textcolor{improve}{(+1.58)}} \\
    \hline
    \multicolumn{1}{l|}{PPL} & \multicolumn{1}{l}{9.93 } & \multicolumn{1}{l}{52.96 } & \multicolumn{1}{l}{97.60 } & \multicolumn{1}{l|}{88.37 } & \multicolumn{1}{l}{7.35 } & \multicolumn{1}{l}{57.54 } & \multicolumn{1}{l}{91.75 } & \multicolumn{1}{l|}{72.65 } & \multicolumn{1}{l}{1.47 } & \multicolumn{1}{l}{53.95 } & \multicolumn{1}{l}{91.99 } & \multicolumn{1}{l}{17.49 } \\
    \multicolumn{1}{l|}{+ MIRA (ours)} & \multicolumn{1}{l}{11.03 \textcolor{improve}{(+1.10)}} & \multicolumn{1}{l}{64.50 \textcolor{improve}{(+11.54)}} & \multicolumn{1}{l}{92.81 \textcolor{improve}{(-4.79)}} & \multicolumn{1}{l|}{92.72 \textcolor{improve}{(+4.35)}} & \multicolumn{1}{l}{5.15 \textcolor{degrade}{(-2.20)}} & \multicolumn{1}{l}{60.81 \textcolor{improve}{(+3.27)}} & \multicolumn{1}{l}{93.09 \textcolor{degrade}{(+1.34)}} & \multicolumn{1}{l|}{79.74 \textcolor{improve}{(+6.91)}} & \multicolumn{1}{l}{1.47 \textcolor{improve}{(+0.00)}} & \multicolumn{1}{l}{58.44 \textcolor{improve}{(+4.49)}} & \multicolumn{1}{l}{88.88 \textcolor{improve}{(-3.11)}} & \multicolumn{1}{l}{46.94 \textcolor{improve}{(+29.45)}} \\
    \hline
    \multicolumn{1}{l|}{Entropy} & \multicolumn{1}{l}{10.66 } & \multicolumn{1}{l}{52.53 } & \multicolumn{1}{l}{98.80 } & \multicolumn{1}{l|}{88.35 } & \multicolumn{1}{l}{8.82 } & \multicolumn{1}{l}{57.41 } & \multicolumn{1}{l}{90.60 } & \multicolumn{1}{l|}{72.63 } & \multicolumn{1}{l}{1.47 } & \multicolumn{1}{l}{53.73 } & \multicolumn{1}{l}{91.33 } & \multicolumn{1}{l}{17.28 } \\
    \multicolumn{1}{l|}{+ MIRA (ours)} & \multicolumn{1}{l}{10.66 \textcolor{improve}{(+0.00)}} & \multicolumn{1}{l}{64.60 \textcolor{improve}{(+12.07)}} & \multicolumn{1}{l}{94.61 \textcolor{improve}{(-4.19)}} & \multicolumn{1}{l|}{91.52 \textcolor{improve}{(+3.17)}} & \multicolumn{1}{l}{6.25 \textcolor{degrade}{(-2.57)}} & \multicolumn{1}{l}{61.11 \textcolor{improve}{(+3.70)}} & \multicolumn{1}{l}{92.51 \textcolor{degrade}{(+1.91)}} & \multicolumn{1}{l|}{74.04 \textcolor{improve}{(+1.41)}} & \multicolumn{1}{l}{1.47 \textcolor{improve}{(+0.00)}} & \multicolumn{1}{l}{57.50 \textcolor{improve}{(+3.77)}} & \multicolumn{1}{l}{92.64 \textcolor{degrade}{(+1.31)}} & \multicolumn{1}{l}{26.02 \textcolor{improve}{(+8.74)}} \\
    \hline
    \multicolumn{1}{l|}{$\Delta$Entropy} & \multicolumn{1}{l}{10.29 } & \multicolumn{1}{l}{49.84 } & \multicolumn{1}{l}{98.80 } & \multicolumn{1}{l|}{88.42 } & \multicolumn{1}{l}{8.09 } & \multicolumn{1}{l}{48.15 } & \multicolumn{1}{l}{96.74 } & \multicolumn{1}{l|}{65.71 } & \multicolumn{1}{l}{1.47 } & \multicolumn{1}{l}{48.20 } & \multicolumn{1}{l}{93.54 } & \multicolumn{1}{l}{14.61 } \\
    \multicolumn{1}{l|}{+ MIRA (ours)} & \multicolumn{1}{l}{10.66 \textcolor{improve}{(+0.37)}} & \multicolumn{1}{l}{60.86 \textcolor{improve}{(+11.02)}} & \multicolumn{1}{l}{92.22 \textcolor{improve}{(-6.58)}} & \multicolumn{1}{l|}{90.20 \textcolor{improve}{(+1.78)}} & \multicolumn{1}{l}{5.51 \textcolor{degrade}{(-2.58)}} & \multicolumn{1}{l}{58.18 \textcolor{improve}{(+10.03)}} & \multicolumn{1}{l}{93.86 \textcolor{improve}{(-2.88)}} & \multicolumn{1}{l|}{70.89 \textcolor{improve}{(+5.18)}} & \multicolumn{1}{l}{2.21 \textcolor{improve}{(+0.74)}} & \multicolumn{1}{l}{60.48 \textcolor{improve}{(+12.28)}} & \multicolumn{1}{l}{89.70 \textcolor{improve}{(-3.84)}} & \multicolumn{1}{l}{19.78 \textcolor{improve}{(+5.17)}} \\
    \hline
    \multicolumn{1}{l|}{CoE-R} & \multicolumn{1}{l}{5.51 } & \multicolumn{1}{l}{51.80 } & \multicolumn{1}{l}{98.20 } & \multicolumn{1}{l|}{88.01 } & \multicolumn{1}{l}{5.51 } & \multicolumn{1}{l}{54.07 } & \multicolumn{1}{l}{89.25 } & \multicolumn{1}{l|}{69.37 } & \multicolumn{1}{l}{3.31 } & \multicolumn{1}{l}{46.99 } & \multicolumn{1}{l}{97.47 } & \multicolumn{1}{l}{17.99 } \\
    \multicolumn{1}{l|}{+ MIRA (ours)} & \multicolumn{1}{l}{12.13 \textcolor{improve}{(+6.62)}} & \multicolumn{1}{l}{59.47 \textcolor{improve}{(+7.67)}} & \multicolumn{1}{l}{93.41 \textcolor{improve}{(-4.79)}} & \multicolumn{1}{l|}{90.02 \textcolor{improve}{(+2.01)}} & \multicolumn{1}{l}{5.51 \textcolor{improve}{(+0.00)}} & \multicolumn{1}{l}{59.84 \textcolor{improve}{(+5.77)}} & \multicolumn{1}{l}{90.60 \textcolor{degrade}{(+1.35)}} & \multicolumn{1}{l|}{73.21 \textcolor{improve}{(+3.84)}} & \multicolumn{1}{l}{0.74 \textcolor{degrade}{(-2.57)}} & \multicolumn{1}{l}{51.49 \textcolor{improve}{(+4.50)}} & \multicolumn{1}{l}{96.30 \textcolor{improve}{(-1.17)}} & \multicolumn{1}{l}{16.32 \textcolor{degrade}{(-1.67)}} \\
    \hline
    \multicolumn{1}{l|}{CoE-C} & \multicolumn{1}{l}{9.93 } & \multicolumn{1}{l}{54.99 } & \multicolumn{1}{l}{97.60 } & \multicolumn{1}{l|}{88.29 } & \multicolumn{1}{l}{6.25 } & \multicolumn{1}{l}{59.65 } & \multicolumn{1}{l}{91.55 } & \multicolumn{1}{l|}{75.17 } & \multicolumn{1}{l}{1.84 } & \multicolumn{1}{l}{59.94 } & \multicolumn{1}{l}{90.27 } & \multicolumn{1}{l}{20.31 } \\
    \multicolumn{1}{l|}{+ MIRA (ours)} & \multicolumn{1}{l}{9.19 \textcolor{degrade}{(-0.74)}} & \multicolumn{1}{l}{63.63 \textcolor{improve}{(+8.62)}} & \multicolumn{1}{l}{89.22 \textcolor{improve}{(-8.38)}} & \multicolumn{1}{l|}{90.64 \textcolor{improve}{(+2.35)}} & \multicolumn{1}{l}{6.62 \textcolor{improve}{(+0.37)}} & \multicolumn{1}{l}{60.47 \textcolor{improve}{(+0.82)}} & \multicolumn{1}{l}{92.51 \textcolor{degrade}{(+0.96)}} & \multicolumn{1}{l|}{71.76 \textcolor{degrade}{(-3.41)}} & \multicolumn{1}{l}{1.84 \textcolor{improve}{(+0.00)}} & \multicolumn{1}{l}{60.68 \textcolor{improve}{(+0.74)}} & \multicolumn{1}{l}{89.21 \textcolor{improve}{(-1.06)}} & \multicolumn{1}{l}{19.09 \textcolor{degrade}{(-1.22)}} \\
    \hline
    Majority\ /\ PRM & \multicolumn{1}{l}{10.66\ /\ 13.97}  & -     & -     & -     & \multicolumn{1}{l}{6.99\ /\ 9.93}   & -     & -     & -     & \multicolumn{1}{l}{3.31\ /\ 6.99}  & -     & -     & -     \\
    \toprule
    \end{tabular}%
}
\end{table*}

\begin{figure}[ht]
    \centering
    \includegraphics[width=\linewidth]{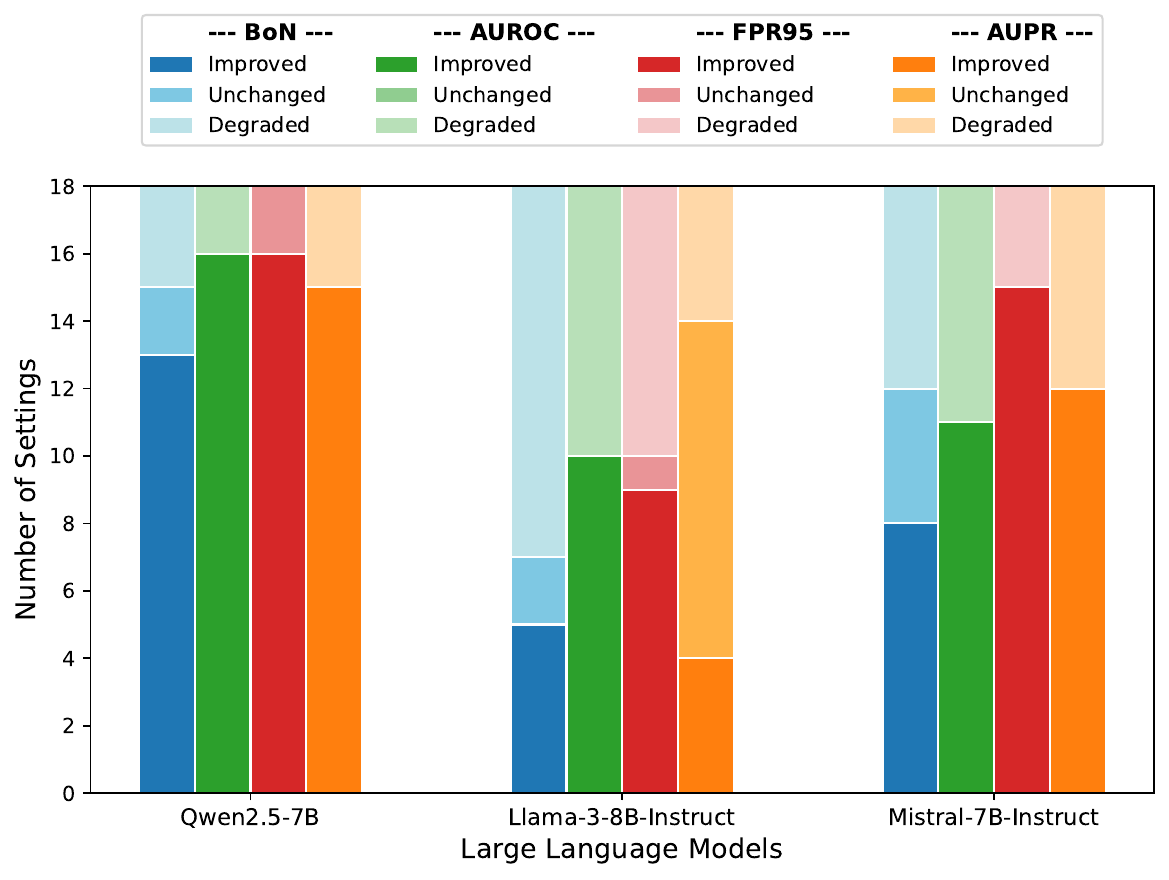}
    \caption{Frequency of MIRA leading to improved, unchanged, and degraded performance for meta-cognition lenses on three LLMs (Qwen2.5-7B, Llama-3-8B-Instruct, Mistral-7B-Instruct) and three datasets (GSM8K, MATH500, MinervaMATH).}
    \label{fig:freq}
\end{figure}

\paragraph{Performance Trends Across Difficulty Levels} 
As illustrated in Table~\ref{fig:comparison}, model performance exhibits an overall correlation with problem difficulty. The aggregate metrics of BoN accuracy, AUROC, and AUPR demonstrate monotonic degradation as task complexity increases. Concurrently, FPR95 shows a statistically significant upward trajectory. Therefore, the performance of the meta-cognition lenses in reflecting reasoning step correctness diminishes substantially on more difficult tasks.

\paragraph{Effectiveness and Efficiency of MIRA} The proposed reward adjust strategy demonstrates robust generalization across model architectures and task difficulties. As detailed in Table~\ref{fig:comparison}, MIRA achieves performance improvements in 61.1\% (BoN) and 68.5\% (AUROC) of experimental configurations (N=54). Figure~\ref{fig:freq} illustrates the particular efficacy of MIRA in enhancing the meta-cognition lenses for Qwen2.5-7B. Although MIRA works on less than half of the settings for Llama-3-8B-Instruct, Table~\ref{exp:comparison} presents that it enhances more meta-cognition lenses on the difficult task, MinervaMATH, than the simple one, GSM8K. Besides, MIRA also performs well on a wide range of meta-cognition lenses for Mistral-7B-Instruct. The gains brought by MIRA across different architectures prove its cross-model compatibility. These experimental results confirm our hypothesis that stepwise adjustment of intrinsic rewards can effectively compensate for inherent calibration weaknesses in the existing lenses of LLM meta-cognition. Although we primarily focus on open-source models, these results are still insightful for closed-source black-box LLMs due to the potential of probability-based meta-cognition lenses such as Maxprob and PPL.
Additionally, Table~\ref{tab:latency-stats} demonstrates that MIRA adds only a negligible amount of latency per instance. The increase does not exceed 0.03 milliseconds, which validates the efficiency of MIRA. 

\begin{table}[htbp]
  \small
  \centering
  \renewcommand{\arraystretch}{1}
  \caption{Latency measurements (unit: millisecond) for the meta-cognition lenses with (w/) and without (w/o) MIRA on Qwen2.5-7B (GSM8K subset). ``Per Step'' and ``Per Instance'' denote the latency averaged across all steps and all instances, respectively.}
  \label{tab:latency-stats}
  {\begin{tabular}{lccc}
    \toprule
    \multirow{2}{*}{\textbf{Methods}} & 
\multirow{2}{*}{\textbf{Per Step}} & \multicolumn{2}{c}{\textbf{\textbf{Per Instance}}}\\
    \cmidrule(lr){3-4}  &  & w/o MIRA & w/ MIRA \\
    \midrule
    Maxprob & 0.12 & 0.64 & 0.65 \\
    \midrule
    PPL & 0.04 & 0.20 & 0.21 \\
    \midrule
    Entropy & 512.94 & 2669.63 & 2669.64 \\
    \midrule
    $\Delta$Entropy & 508.54 & 2646.73 & 2646.74 \\
    \midrule
    CoE-R & 0.51 & 2.64 & 2.66 \\
    \midrule
    CoE-C & 0.64 & 3.31 & 3.34 \\
    \bottomrule
    \end{tabular}
    }
\end{table}
\paragraph{Validation on Difficult Math Tasks} Practical validation on the MinervaMATH dataset reveals compelling advantages of meta-cognition lenses over conventional ensemble approaches. For instance, Qwen2.5-7B with adjusted CoE-R configuration outperforms majority voting baselines by 1.47\% in BoN accuracy (12.13\% vs 10.66\%), while Llama-3-8B-Instruct achieves better performance through self-evaluation-based selection than majority voting (PPL: +0.36\%, Entropy: +1.83\%, $\Delta$entropy: +1.10\%). This performance gap suggests that properly calibrated self-evaluation metrics enable more effective identification of high-quality reasoning paths than static aggregation methods. The findings align with our hypothesis that LLMs contain latent self-diagnostic capabilities that can be operationalized through appropriate metric design. Meanwhile, these results also indicate the potential of meta-cognition evaluation as a mechanism for autonomous LLM self-improvement through intrinsic process supervision.

\subsection{PRM-as-a-Judge Analysis}
\label{exp:prm}
To validate the reasonableness of utilizing PRM as an annotator, we conduct the following analysis: (1) Analyze the consistency between the rewards of two different PRMs, Qwen2.5-Math-PRM-7B and Skywork-o1-Open-PRM-Qwen-2.5-7B~\cite{skyworkopeno12024}; (2) Compare AutoMeco with BoN by calculating the consistency of their benchmarking results of six meta-cognition lenses. The consistency metrics include top-K and last-K match rate, and consistency rate (CR)~\cite{wang2025cream}. Appendix~\ref{consistency} presents more details on these metrics.

\paragraph{Agreement between Different PRMs} 
Table~\ref{tab:annotation_consistency} shows strong inter-annotator agreement between the two PRMs, confirming their reliability for evaluating step-level errors. Notably, the kappa score for annotations on Mistral-7B-Instruct's reasoning traces reaches 0.7334. However, the PRMs exhibit lower agreement on step-level annotations than on instance-level ones. For example, the step-level Kappa for the reasoning steps of Qwen2.5-7B is 0.3274, compared to 0.5878 at the instance level. This discrepancy indicates that while the PRMs are effective judges overall, there is room to improve their fine-grained annotation accuracy.

\begin{table}[h!]
\small
\centering
\renewcommand{\arraystretch}{1.3}
\caption{Inter-annotator agreement for step- and instance-level rewards between Qwen2.5-Math-PRM-7B and Skywork-o1-Open-PRM-Qwen-2.5-7B on the responses of the three LLMs to the three math datasets. ``Spearman'' and ``Pearson'' are the Spearman~\cite {spearman1961proof} and Pearson's correlation coefficients~\cite{pearson1895vii}. All the correlations in this table are statistically significant (p < 0.05). ``Cohen'' represents the inter-annotator agreement score, Cohen's Kappa~\cite{mchugh2012interrater}. The reported Kappa score is maximized by finding the optimal binarization thresholds through a grid search over the range \{0.1, 0.2, $\cdots$, 0.8, 0.9\}.}
\label{tab:annotation_consistency}
\resizebox{\linewidth}{!}{\begin{tabular}{@{}cccc@{}}
\toprule
\textbf{Model} & \textbf{Spearman}$\uparrow$ & \textbf{Pearson}$\uparrow$& \textbf{Cohen}$\uparrow$\\
\hline
\rowcolor{gray!20}\multicolumn{4}{c}{Step Level} \\
\hline
Qwen2.5-7B & 0.6309 & 0.3552  & 0.3274  \\
Llama-3-8B-Instruct & 0.5850 & 0.3473 & 0.2657 \\
Mistral-7B-Instruct & 0.5758 & 0.6109 & 0.5349 \\
\hline
\rowcolor{gray!20}\multicolumn{4}{c}{Instance Level (Averaged on Steps)} \\
\hline
Qwen2.5-7B  & 0.7630  & 0.6236  & 0.5878 \\
Llama-3-8B-Instruct & 0.7179 & 0.6617 & 0.6108 \\
Mistral-7B-Instruct & 0.7250 & 0.7981 & 0.7334\\
\bottomrule
\end{tabular}
}
\end{table}
\paragraph{Consistency between AutoMeco and BoN} Our analysis demonstrates that PRM-as-a-Judge achieves reasonable consistency with BoN evaluation, validating its utility as a complementary method for efficient method ranking. While perfect alignment is not observed, the results highlight meaningful agreement trends.
As shown in Table~\ref{tab:consistency}, AutoMeco exhibits alignment with BoN rankings for top-tier methods, particularly at K=3. BoN’s top methods include its top three in 66.67\% settings on average, with Mistral-7B-Instruct achieving 100\% consistency. Furthermore, the average CR of 48.15\% supports the reliability of AutoMeco to benchmark the meta-cognition lenses. Moreover, it achieves the highest alignment on Llama-3-8B-Instruct (CR=53.33\%), suggesting model-specific optimization potential.

\begin{table}[!h]
\centering
\renewcommand\arraystretch{1.3}
\caption{Consistency metrics of Best-of-N (BoN) and AutoMeco results across three large language models. Top-k and Last-K Match evaluate whether the best/worst method chosen by AutoMeco is in the top/last K methods ranked by BoN. Top-K Order considers both the best and the worst. CR stands for consistency rate.}
\label{tab:consistency}
\sisetup{detect-weight, mode=text}
\resizebox{\linewidth}{!}{\begin{tabular}{cccccccc}
\toprule
\multirow{2.5}{*}{\textbf{Model}} & 
\multicolumn{2}{c}{\textbf{Top-K Match}} & \multicolumn{2}{c}{\textbf{Last-K Match}} & \multicolumn{2}{c}{\textbf{Top-K Order}} & \multirow{2}{*}{\textbf{CR}} \\
\cmidrule(lr){2-3} \cmidrule(lr){4-5} \cmidrule(lr){6-7}
& {K=1}  & {K=3} & {K=1} & {K=3} & {K=1} & {K=3} & \\
\hline
Qwen2.5  & 0.00  & 66.67 & 0.00 & 66.67 & 0.00  & 16.67 & 44.44 \\
Llama-3 & 0.00  & 33.33 & 0.00 & 33.33 & 0.00  & 0.00 & 53.33 \\
Mistral & 33.33  & 100.00 & 0.00  & 33.33 & 0.00  & 16.67 & 46.67 \\
\hline
Average & 11.11 & 66.67 & 0.00 & 44.44 & 0.00  & 11.11 & 48.15 \\
\bottomrule
\end{tabular}
}
\end{table}

\section{Conclusion}
\label{conclusion}
We investigate the LLM meta-cognition observation capability of self-evaluation measures as meta-cognition lenses for language models, through an automated benchmarking framework to evaluate these lenses and a fine-grained self-evaluation adjustment strategy to enhance them. These meta-cognition lenses can capture the LLM meta-cognition, and the stepwise modification further improves their observation ability.

Our study points to several directions with a considerable range of research for future work, including constructing an LLM meta-cognition benchmark with manually annotated step correctness and the LLM internal states for errorless step labels, developing more accurate self-evaluation measures for meta-cognition observation, applying meta-cognition signals to realize LLM self-improvement, and aligning LLM with human preferences more efficiently via meta-cognition loss. Moreover, utilizing the meta-cognition lenses for response refinement in other scenarios, such as agentic tasks, constitutes a crucial direction.

\section*{Limitations}
Our work focuses on introducing meta-cognition into language model evaluation by incorporating internal model states into self-assessment mechanisms. While our approach demonstrates promising results, several limitations warrant discussion:

\paragraph{Model Accessibility Requirements} 
Our method incorporates internal model representations such as hidden states to enable fine-grained self-evaluation. This design, however, requires access to model internals and is mainly limited to open-source architectures. While logits- and probability-based self-evaluation methods are compatible with both open-source and mainstream closed-source LLMs with access to the last-layer states, more delicate approaches using hidden states cannot be directly applied to closed-source models such as GPT-4~\cite{achiam2023gpt}. Exploring approximations or hybrid strategies may help bridge the gap between white-box and black-box interpretability.

\paragraph{Computational and Memory Overhead}
Our Markovian Intrinsic Reward Adjustment (MIRA) strategy utilizes internal model signals to dynamically refine self-evaluation. As a result, it introduces additional computation during the reasoning phase and requires extra memory to store intermediate hidden states. Though the overhead remains moderate in scale, these factors may influence efficiency in practical deployment settings.

\paragraph{Extension to Large Reasoning Models}  
Our framework has been validated across multiple model families, including Qwen, Llama, and Mistral. However, generalizing it to more sophisticated Large Reasoning Models (LRMs) involves non-trivial considerations. 
The dynamic and complex nature of their reasoning processes calls for more adaptive step-wise modeling techniques. Integrating these with recent advances in cognitive behavior analysis of RL~\cite{yue2025does, gandhi2025cognitive} or reasoning boundary analysis~\cite{chen2024unlocking} may offer a promising path forward.


\section*{Acknowledgments}
The authors would like to thank all of the anonymous reviewers for their valuable comments. This research is funded by the National Natural Science Foundation of China (Grant No.62176053). This work is supported by the Big Data Computing Center of Southeast University.

\bibliography{custom}

\newpage
\clearpage
\appendix

\section{LLM Self-Evaluation Measures}
\label{sec:self-evaluation-methods}
\subsection{Chain-of-Embedding (CoE)}
For a reasoning step $R_i$ with $T_i$ tokens, \citet{wang2025latent} quantifies two features, Magnitude and Angle, to represent the layer-by-layer changes of query understanding based on hidden states:
\begin{align*}
    \textrm{Mag}(\boldsymbol{H}_i) &= \frac{1}{L}\sum_{\ell=0}^{L-1}\frac{||\boldsymbol{h}_{\ell+1}-\boldsymbol{h}_\ell||_2}{||\boldsymbol{h}_{L}-\boldsymbol{h}_0||_2}\\
    \textrm{Ang}(\boldsymbol{H}_i) &= \frac{1}{L}\sum_{\ell=0}^{L-1}\frac{\textrm{arccos}\left(\frac{\boldsymbol{h}_{\ell+1}^\intercal\ \cdot\ \boldsymbol{h}_{\ell}}{||\boldsymbol{h}_{\ell+1}||_2\ \cdot \ ||\boldsymbol{h}_{\ell}||_2}\right)}{\textrm{arccos}\left(\frac{\boldsymbol{h}_{L}^\intercal\ \cdot\ \boldsymbol{h}_{0}}{||\boldsymbol{h}_{L}||_2\ \cdot \ ||\boldsymbol{h}_{0}||_2}\right)}
\end{align*}

\begin{equation}
    \boldsymbol{h}^\ell=\frac{1}{T_i}\sum_{t=1}^{T_i} \boldsymbol{h}_t^\ell,\ \forall\ \ell\in [0,...,L]\nonumber
\end{equation}

Two basic components in Magnitude and Angle, magnitude change $M(\boldsymbol{h}_\ell,\boldsymbol{h}_{\ell+1})$ and angle change $A(\boldsymbol{h}_\ell,\boldsymbol{h}_{\ell+1})$, are defined as follows:

\begin{align*}
    M(\boldsymbol{h}_\ell,\boldsymbol{h}_{\ell+1})&=||\boldsymbol{h}_{\ell+1}-\boldsymbol{h}_\ell||_2\\
    A(\boldsymbol{h}_\ell,\boldsymbol{h}_{\ell+1}) &= \textrm{arccos}\left(\frac{\boldsymbol{h}_{\ell+1}^\intercal\ \cdot\ \boldsymbol{h}_{\ell}}{||\boldsymbol{h}_{\ell+1}||_2\ \cdot \ ||\boldsymbol{h}_{\ell}||_2}\right)
\end{align*}

As shown in Equation~\ref{eq:coe-r} and~\ref{coe-c}, CoE-R and CoE-C capture the correctness by combining the magnitude and angle changes in the real and complex spaces, respectively~\cite{wang2025latent}.

\begin{figure*}
\vspace{-10pt}
\small
    \begin{equation}
    \textrm{CoE-R}(\boldsymbol{H}_i) =  \ \textrm{Mag}(\boldsymbol{H}_i)-\textrm{Ang}(\boldsymbol{H}_i)
    =\frac{1}{L}\sum_{\ell=0}^{L-1}\left(\frac{M(\boldsymbol{h}_\ell,\boldsymbol{h}_{\ell+1})}{M(\boldsymbol{h}_0,\boldsymbol{h}_{L})}-\frac{A(\boldsymbol{h}_\ell,\boldsymbol{h}_{\ell+1})}{A(\boldsymbol{h}_0,\boldsymbol{h}_{L})}\right)
    \label{eq:coe-r}
    \end{equation}
\small
    \begin{align}
    \textrm{CoE-C}(\boldsymbol{H}_i)&=\frac{1}{L}\left| \sum_{\ell=0}^{L-1}M(\boldsymbol{h}_\ell,\boldsymbol{h}_{\ell+1})\cdot e^{i\cdot A(\boldsymbol{h}_\ell,\boldsymbol{h}_{\ell+1})}\right|\nonumber\\
    &=\frac{1}{L} \left| \sum_{\ell=0}^{L-1} \left[M(\boldsymbol{h}_\ell,\boldsymbol{h}_{\ell+1})\cos(A(\boldsymbol{h}_\ell,\boldsymbol{h}_{\ell+1}))+i\cdot M(\boldsymbol{h}_\ell,\boldsymbol{h}_{\ell+1})\sin(A(\boldsymbol{h}_\ell,\boldsymbol{h}_{\ell+1})\right]
    \right|\nonumber\\
    &=\frac{1}{L}\sqrt{\left(\sum_{\ell=0}^{L-1}M(\boldsymbol{h}_\ell,\boldsymbol{h}_{\ell+1})\cos(A(\boldsymbol{h}_\ell,\boldsymbol{h}_{\ell+1})) \right)^2+\left(\sum_{\ell=0}^{L-1}M(\boldsymbol{h}_\ell,\boldsymbol{h}_{\ell+1})\sin(A(\boldsymbol{h}_\ell,\boldsymbol{h}_{\ell+1})) \right)^2}\label{coe-c}
\end{align}
\end{figure*}

\subsection{Entropy}
Entropy reflects the uncertainty of reasoning steps based on their token probabilities~\cite{si2022prompting}:
\begin{equation}
    \textrm{Entropy}(\boldsymbol{P}_i)=\frac{1}{T_i}\sum_{t=1}^{T_i}\left(-\boldsymbol{p}_t^\intercal\cdot\textrm{log}\ \boldsymbol{p}_t\right)
\end{equation}

We utilize its reciprocal as the correctness score.

\subsection{$\Delta$Entropy}
\citet{yin-etal-2024-reasoning} proposes that the abnormal uncertainty fluctuation is useful for judging wrong steps during reasoning, which is formally the uncertainty change between two adjacent reasoning steps. The more the LLM's uncertainty fluctuates, the more likely the LLM makes mistakes. We choose Entropy as the uncertainty metric to formulate the entropy fluctuation:

\begin{footnotesize}
\begin{equation}
    \Delta\textrm{Entropy}(\boldsymbol{P}_i) = 
    \begin{cases}
        0,\ \textrm{if} \ i=0,\\
        \textrm{Entropy}(\boldsymbol{P}_i)-\textrm{Entropy}(\boldsymbol{P}_{i-1}), \ \textrm{else.}
    \end{cases}
\end{equation}
\end{footnotesize}

We utilize the opposite number of $\Delta$Entropy as the step correctness score.

\subsection{Max Probability (Maxprob)}
Maxprob calculates the average of the maximal elements in the probability distributions, which assumes that the top-1 probability reflects the certainty of LLMs:
\begin{equation}
    \textrm{Maxprob}(\boldsymbol{P}_i) = \frac{1}{T_i}\sum_{t=1}^{T_i}\max(\boldsymbol{p}_t)
\end{equation}

\subsection{Perplexity (PPL)}
PPL reflects the uncertainty of LLMs by considering the negative logarithm of the maximal probability:
\begin{equation}
    \textrm{PPL}(\boldsymbol{P}_i)=-\frac{1}{T_i}\sum_{t=1}^{T_i}\ \textrm{log} \ \max(\boldsymbol{p}_t)
\end{equation}

We utilize its reciprocal as the correctness score.
  
\section{Consistency Metrics}
\label{consistency}
For an LLM $\Theta$ and a dataset $\mathcal{D}$, AutoMeco and BoN evaluate $M$ meta-cognition lenses, which results in two ranks denoted by $\boldsymbol{\alpha}=\alpha_1, \alpha_2,...,\alpha_M$ and $\boldsymbol{\beta}=\beta_1, \beta_2,...,\beta_M$. Top-K Match, Last-K Match, and Top-K Order are measured as follows:

\begin{footnotesize}
\begin{align*}
\textrm{TopMatch}_{\Theta,\mathcal{D}}(K)&=\mathbb{I}[\textrm{argmin}\ \alpha_m \in \textrm{argsort}(\boldsymbol{\beta})[:K]]\\
\textrm{LastMatch}_{\Theta,\mathcal{D}}(K)&=\mathbb{I}[\textrm{argmax}\ \alpha_m \in \textrm{argsort}(\boldsymbol{\beta})[-K:]]\\
\textrm{TopOrder}_{\Theta,\mathcal{D}}(K)&=\textrm{TopMatch}_{\Theta,\mathcal{D}}(K)\cdot\textrm{LastMatch}_{\Theta,\mathcal{D}}(K)
\end{align*}
\end{footnotesize}

\noindent where $K\in [1,2,3]$,  $\mathbb{I}[\cdot]$ is the indicator function and $\textrm{argsort}(\boldsymbol{\beta})$ outputs the indices that sort $\boldsymbol{\beta}$ in ascending order.
Equation~\ref{cr} calculates the consistency rate of $\boldsymbol{\alpha}$ and $\boldsymbol{\beta}$~\cite{wang2025cream}.

\begin{align}\label{cr}
    \textrm{CR}_{\Theta,\mathcal{D}} = &\dfrac{2}{M(M-1)}\sum_{1\leq m<m'\leq M}\nonumber\\
    [&\mathbb{I}[(\alpha_m-\alpha_{m'})(\beta_m-\beta_{m'})>0]-\nonumber\\
    &\mathbb{I}[(\alpha_m-\alpha_{m'})(\beta_m-\beta_{m'})<0]]
\end{align}

Ultimately, we average these metrics on three datasets to present the consistency
of AutoMeco and BoN for the LLM $\Theta$.

\section{Prompt Templates}
\label{sec:prompt}
\subsection*{\textbullet\  GSM8K}
\begin{problemformat}

\begin{itemize}[leftmargin=0em]
    \item[] Solve this math problem step by step. Give the reasoning steps using `Step $n$:' before each step to distinguish between different steps, where $n$ is a positive integer starting from 1, representing the current step number. Then give the final answer on the last line by itself in the format of \textcolor{red}{\texttt{"Answer:"}}
    \item[] Do not add anything other than the integer answer after \textcolor{red}{\texttt{``Answer:''}}
    \item[] \textbf{Question:} \texttt{\{input\_data\}}
\end{itemize}
\end{problemformat}

\subsection*{\textbullet\ MATH500 and MinervaMATH}
\begin{problemformat}
\begin{itemize}[leftmargin=0em]
    \item[] \textbf{Question:} \texttt{\{input\_data\}}
    \item[] Please reason step by step. Use `Step $n$:' before each step to distinguish between different steps, where $n$ is a positive integer starting from 1, representing the current step number. Then, give your final answer on the last line in the format of \textcolor{red}
    {\texttt{``Answer:}}
    \textcolor{blue}{\texttt{\textbackslash boxed\{\}}} 
    \textcolor{red} {\texttt{''}}

\end{itemize}
\end{problemformat}

    
    

\section{Statistical Feasibility on Other LLMs}
As shown in Table~\ref{tab:llama-correlation} and \ref{tab:mistral-correlation}, meta-cognition lenses perform weakly in capturing meta-cognition of Llama-3-8B-Instruct and Mistral-7B-Instruct. Interestingly, however, these methods are statistically better at hard tasks than simple ones. For instance, CoE-C has a stronger correlation with PRM rewards on MinervaMATH than GSM8K and MATH500 for the two LLMs.

\begin{table*}[h]
\centering
\renewcommand\arraystretch{1.6}
\caption{Spearman coefficient~\cite{spearman1961proof} and Kendall's Tau~\cite{kendall1938new} between the intrinsic and PRM rewards of Llama-3-8B-Instruct on three datasets.}
\label{tab:llama-correlation}
\begin{tabular}{ccccccc}
\toprule
\multirow{2}{*}{\textbf{Methods}} & \multicolumn{2}{c}{\textbf{GSM8K}} & \multicolumn{2}{c}{\textbf{MATH500}} & \multicolumn{2}{c}{\textbf{MinervaMATH}} \\
\cmidrule(lr){2-3} \cmidrule(lr){4-5} \cmidrule(lr){6-7}
  & \textit{Spearman} & \textit{Kendall} & \textit{Spearman} & \textit{Kendall} & \textit{Spearman} & \textit{Kendall} \\
\hline
\texttt{CoE-C} & \makecell[c]{0.179 \\ {\footnotesize (8.0e-10)}} 
& \makecell[c]{0.119 \\ {\footnotesize (1.1e-9)}} 
& \makecell[c]{0.095 \\ {\footnotesize (4.0e-8)}} 
& \makecell[c]{0.054 \\ {\footnotesize (3.1e-6)}} 
& \makecell[c]{\underline{0.203} \\ {\footnotesize (8.2e-15)}} 
& \makecell[c]{\underline{0.135} \\ {\footnotesize (2.0e-14)}} \\
\hline
\texttt{CoE-R} & \makecell[c]{\textbf{0.219} \\ {\footnotesize (3.8e-14)}} 
& \makecell[c]{\textbf{0.139} \\ {\footnotesize (1.4e-12)}}  
& \makecell[c]{\textbf{0.123} \\ {\footnotesize (1.2e-12)}} 
& \makecell[c]{\textbf{0.083} \\ {\footnotesize (9.2e-13)}} 
& \makecell[c]{0.155 \\ {\footnotesize (3.8e-9)}} 
& \makecell[c]{0.103 \\ {\footnotesize (4.5e-9)}} \\
\hline
\texttt{Maxprob} & \makecell[c]{0.102 \\ {\footnotesize (0.0005)}} 
& \makecell[c]{0.081 \\ {\footnotesize (3.6e-5)}} 
& \makecell[c]{0.098 \\ {\footnotesize (1.5e-8)}} 
& \makecell[c]{0.063 \\ {\footnotesize (6.9e-8)}} 
& \makecell[c]{0.022 \\ {\footnotesize (0.4058)}} 
& \makecell[c]{0.013 \\ {\footnotesize (0.4569)}} \\
\hline
\texttt{PPL} & \makecell[c]{0.100 \\ {\footnotesize (0.0006)}} 
& \makecell[c]{0.080 \\ {\footnotesize (4.2e-5)}} 
& \makecell[c]{0.101 \\ {\footnotesize (6.1e-9)}} 
& \makecell[c]{0.064 \\ {\footnotesize (3.4e-8)}} 
& \makecell[c]{\textbf{0.263} \\ {\footnotesize (2.2e-22)}} 
& \makecell[c]{\textbf{0.180} \\ {\footnotesize (9.1e-23)}} \\
\hline
\texttt{Entropy} & \makecell[c]{0.119 \\ {\footnotesize (5.0e-5)}} 
& \makecell[c]{0.092 \\ {\footnotesize (2.3e-6)}} 
& \makecell[c]{\underline{0.104} \\ {\footnotesize (2.0e-9)}} 
& \makecell[c]{\underline{0.065} \\ {\footnotesize (2.6e-8)}} 
& \makecell[c]{0.044 \\ {\footnotesize (0.098)}} 
& \makecell[c]{0.027 \\ {\footnotesize (0.127)}} \\
\hline
\texttt{$\Delta$Entropy} & \makecell[c]{ \underline{-0.208}  \\ {\footnotesize (8.0e-13)}} 
& \makecell[c]{\underline{-0.136} \\ {\footnotesize (3.4e-12)}} 
& \makecell[c]{0.024 \\ {\footnotesize (0.167)}} 
& \makecell[c]{0.019 \\ {\footnotesize (0.094)}} 
& \makecell[c]{-0.074 \\ {\footnotesize (0.0047)}} 
& \makecell[c]{-0.049 \\ {\footnotesize  ( 0.00499)}} \\
\bottomrule
\end{tabular}
\end{table*}

\begin{table*}[h]
\centering
\renewcommand\arraystretch{1.6}
\caption{Spearman coefficient~\cite{spearman1961proof} and Kendall's Tau~\cite{kendall1938new} between the intrinsic and PRM rewards of Mistral-7B-Instruct on three datasets.}
\label{tab:mistral-correlation}
\begin{tabular}{ccccccc}
\toprule
\multirow{2}{*}{\textbf{Methods}} & \multicolumn{2}{c}{\textbf{GSM8K}} & \multicolumn{2}{c}{\textbf{MATH500}} & \multicolumn{2}{c}{\textbf{MinervaMATH}} \\
\cmidrule(lr){2-3} \cmidrule(lr){4-5} \cmidrule(lr){6-7}
  & \textit{Spearman} & \textit{Kendall} & \textit{Spearman} & \textit{Kendall} & \textit{Spearman} & \textit{Kendall} \\
\hline
\texttt{CoE-C} & \makecell[c]{0.194 \\ {\footnotesize (9.0e-08)}} 
& \makecell[c]{0.130 \\ {\footnotesize (1.1e-07)}} 
& \makecell[c]{\textbf{0.215} \\ {\footnotesize (3.9e-17)}} 
& \makecell[c]{\textbf{0.139} \\ {\footnotesize (5.9e-16)}} 
& \makecell[c]{\textbf{0.333} \\ {\footnotesize (2.5e-18)}} 
& \makecell[c]{\textbf{0.221} \\ {\footnotesize (2.7e-17)}} \\
\hline
\texttt{CoE-R} & \makecell[c]{-0.282 \\ {\footnotesize (3.9e-15)}} 
& \makecell[c]{-0.190 \\ {\footnotesize (9.1e-15)}}  
& \makecell[c]{\underline{-0.118} \\ {\footnotesize (4.8e-06)}} 
& \makecell[c]{\underline{-0.078} \\ {\footnotesize (5.8e-06)}} 
& \makecell[c]{\underline{-0.101} \\ {\footnotesize (0.0099)}} 
& \makecell[c]{\underline{-0.064} \\ {\footnotesize (0.0148)}} \\
\hline
\texttt{Maxprob} & \makecell[c]{0.306 \\ {\footnotesize (1.1e-17)}} 
& \makecell[c]{0.203 \\ {\footnotesize (1.1e-16)}} 
& \makecell[c]{0.036 \\ {\footnotesize (0.158)}} 
& \makecell[c]{0.023 \\ {\footnotesize (0.185)}} 
& \makecell[c]{0.011 \\ {\footnotesize (0.783)}} 
& \makecell[c]{0.005 \\ {\footnotesize (0.848)}} \\
\hline
\texttt{PPL} & \makecell[c]{\underline{0.320} \\ {\footnotesize (3.1e-19)}} 
& \makecell[c]{\underline{0.213} \\ {\footnotesize (3.7e-18)}} 
& \makecell[c]{0.051 \\ {\footnotesize (0.048)}} 
& \makecell[c]{0.032 \\ {\footnotesize (0.061)}} 
& \makecell[c]{0.033 \\ {\footnotesize (0.396)}} 
& \makecell[c]{0.020 \\ {\footnotesize (0.437)}} \\
\hline
\texttt{Entropy} & \makecell[c]{\textbf{0.326} \\ {\footnotesize (6.8e-20)}} 
& \makecell[c]{\textbf{0.217} \\ {\footnotesize (7.2e-19)}} 
& \makecell[c]{0.035 \\ {\footnotesize (0.181)}} 
& \makecell[c]{0.023 \\ {\footnotesize (0.191)}} 
& \makecell[c]{0.012 \\ {\footnotesize (0.754)}} 
& \makecell[c]{0.007 \\ {\footnotesize (0.794)}} \\
\hline
\texttt{$\Delta$Entropy} & \makecell[c]{0.154 \\ {\footnotesize (2.4e-05)}} 
& \makecell[c]{0.105 \\ {\footnotesize (2.1e-05)}} 
& \makecell[c]{0.009 \\ {\footnotesize (0.713)}} 
& \makecell[c]{0.006 \\ {\footnotesize (0.721)}} 
& \makecell[c]{-0.040 \\ {\footnotesize (0.303)}} 
& \makecell[c]{-0.027 \\ {\footnotesize (0.301)}} \\
\bottomrule
\end{tabular}
\end{table*}

\section{Statistics of Reasoning Steps}

\begin{table}[H]
\centering
\renewcommand\arraystretch{1.07}
\caption{Statistics on correct, wrong, and uncertain reasoning steps annotated by Qwen2.5-Math-PRM-7B in mathematical reasoning tasks.}
\label{tab:reasoning-steps}
\resizebox{\linewidth}{!}{\begin{tabular}{llrrr}
\toprule
\multirow{2}{*}{\textbf{Model}} & \multirow{2}{*}{\textbf{Dataset}} & \multicolumn{3}{c}{\textbf{Step Correctness}} \\
\cmidrule(lr){3-5}
 & & Correct & Wrong & Uncertain \\
\midrule
\multirow{3}{*}{Qwen2.5} 
 & GSM8K & 1127 & 17 & 122 \\
 & MATH500 & 3079 & 106 & 471 \\
 & MinervaMATH & 1278 & 47 & 277 \\
\hline

\multirow{3}{*}{Llama-3}
 & GSM8K & 1130 & 34 & 229 \\
 & MATH500 & 2900 & 401 & 1135 \\
 & MinervaMATH & 1328 & 113 & 637 \\
\hline

\multirow{3}{*}{Mistral}
 & GSM8K & 523 & 223 & 828 \\
 & MATH500 & 3079 & 106 & 471 \\
 & MinervaMATH & 371 & 1130 & 2322 \\
\bottomrule
\end{tabular}}
\end{table}

\section{Case Study}
\begin{figure*}[b]
    \centering
    \includegraphics[width=0.9\linewidth]{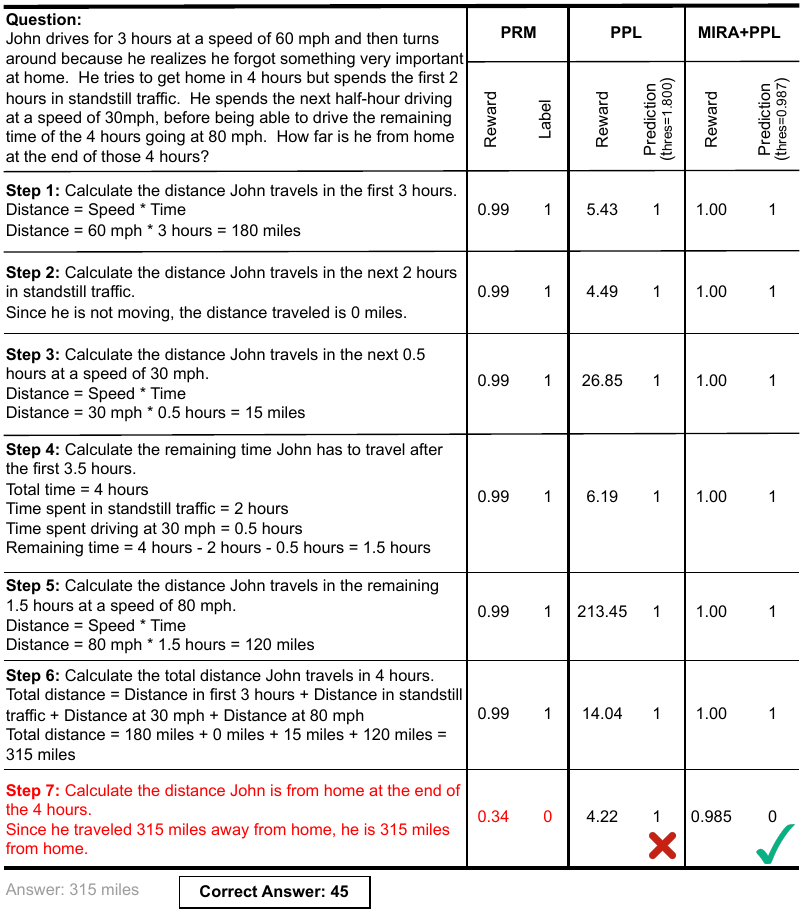}
    \caption{A demonstration of MIRA enhancing meta-cognition observation for Qwen2.5-7B on GSM8K. The thresholds are determined by maximizing the F1 score of step correctness prediction.}
    \label{fig:gsm8k-demo}
\end{figure*}

\begin{figure*}[b]
    \centering
    \includegraphics[width=0.9\linewidth]{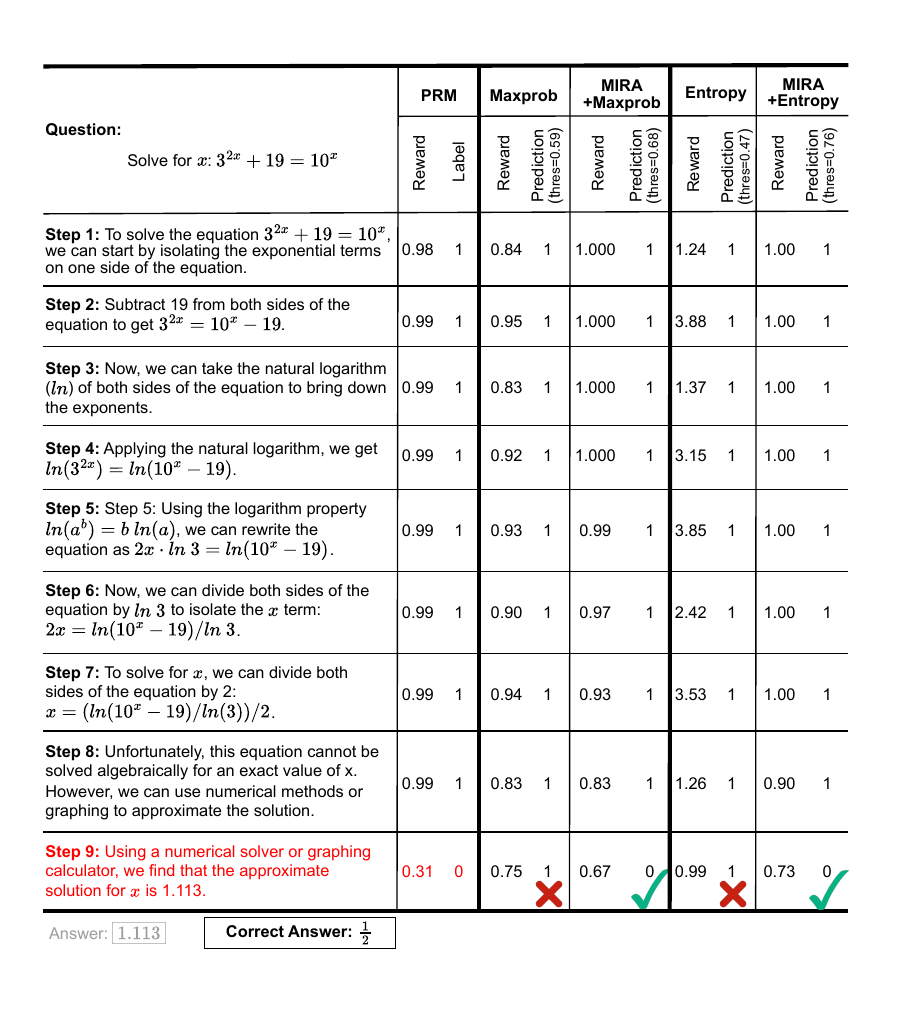}
    \caption{A demonstration of MIRA enhancing meta-cognition observation for Qwen2.5-7B on MATH500. The thresholds are determined by maximizing the F1 score of step correctness prediction. }
    \label{fig:math500-demo}
\end{figure*}

\begin{figure*}[b]
    \centering
    \includegraphics[width=0.9\linewidth]{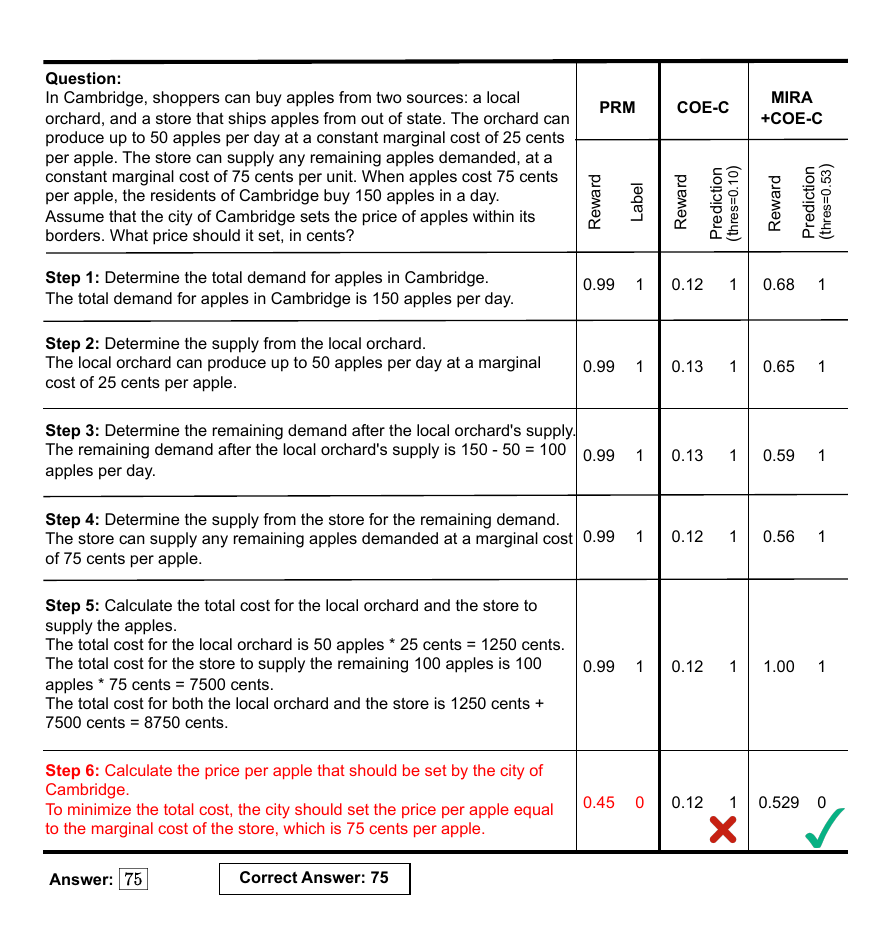}
    \caption{A demonstration of MIRA enhancing meta-cognition observation for Qwen2.5-7B on MinervaMATH. The thresholds are determined by maximizing the F1 score of step correctness prediction.}
    \label{fig:minerva-demo}
\end{figure*}

Figure~\ref{fig:gsm8k-demo}, \ref{fig:math500-demo}, and \ref{fig:minerva-demo} illustrate three cases of Qwen2.5-7B on GSM8K, MATH500, and MinervaMATH datasets. In these cases, self-evaluation measures fail to predict the rightness of the wrong step, while MIRA successfully calibrates the intrinsic meta-cognition by conducting stepwise adjustments to these meta-cognition lenses.

\end{document}